\documentclass[default,iicol,sn-mathphys]{sn-jnl}

\jyear{2022}%
\usepackage{graphicx}
\usepackage{subfigure}
\usepackage{amsmath}
\usepackage{multirow}
\usepackage{color}
\usepackage{caption}
\usepackage{xspace}
\usepackage{amsfonts,amssymb}

\makeatletter
\DeclareRobustCommand\onedot{\futurelet\@let@token\@onedot}
\def\@onedot{\ifx\@let@token.\else.\null\fi\xspace}

\def\eg{\emph{e.g}\onedot} 
\def\ie{\emph{i.e}\onedot} 
 
\def\etc{\emph{etc}\onedot} 
 
\def\etal{\emph{et al}\onedot}
\makeatother

\theoremstyle{thmstyleone}%
\theoremstyle{thmstyletwo}%

\theoremstyle{thmstylethree}%

\begin{document}

\title[RealDAN]{End-to-end Alternating Optimization for Real-World Blind Super Resolution}


\author[1,2,3]{\sur{Zhengxiong Luo}}\email{zhengxiong.luo@cripac.ia.ac.cn}

\author*[2,3]{\sur{Yan Huang}}\email{yhuang@nlpr.ia.ac.cn}

\author[1,2]{\sur{Shang Li}}\email{lishang2018@ia.ac.cn}

\author[2,3,4]{\sur{Liang Wang}}\email{wangliang@nlpr.ia.ac.cn}

\author[2,3,5]{\sur{Tieniu Tan}}\email{tnt@nlpr.ia.ac.cn}

\affil[1]{\orgdiv{Artificial Intelligence School}, \orgname{University of Chinese Academy of Sciences}, \city{Beijing}, \postcode{100049}, \country{China}}

\affil[2]{\orgdiv{Institute of Automation}, \orgname{Chinese Academy of Sciences}, \city{Beijing}, \postcode{100190}, \country{China}}

\affil[3]{\orgname{Center for Research on Intelligent Perception and Computing}, \city{Beijing}, \country{China}}

\affil[4]{\orgname{Center for Excellence in Brain Science and Intelligence Technology}, \city{Beijing}, \country{China}}

\affil[5]{
	\orgname{Nanjing University}, \city{Nanjing}, \country{China}
}


\abstract{
	Blind Super-Resolution (SR) usually involves two sub-problems: 1) estimating the degradation of the given low-resolution (LR) image; 2) super-resolving the LR image to its high-resolution (HR) counterpart. Both problems are ill-posed due to the information loss in the degrading process. Most previous methods try to solve the two problems independently, but often fall into a dilemma: a good super-resolved HR result requires an accurate degradation estimation, which however, is difficult to be obtained without the help of original HR information. To address this issue, instead of considering these two problems independently, we adopt an alternating optimization algorithm, which can estimate the degradation and restore the SR image in a single model. Specifically, we design two convolutional neural modules, namely \textit{Restorer} and \textit{Estimator}. \textit{Restorer} restores the SR image based on the estimated degradation, and \textit{Estimator} estimates the degradation with the help of the restored SR image. We alternate these two modules repeatedly and unfold this process to form an end-to-end trainable network. In this way, both \textit{Restorer} and \textit{Estimator} could get benefited from the intermediate results of each other, and make each sub-problem easier. Moreover, \textit{Restorer} and \textit{Estimator} are optimized in an end-to-end manner, thus they could get more tolerant of the estimation deviations of each other and cooperate better to achieve more robust and accurate final results. Extensive experiments on both synthetic datasets and real-world images show that the proposed method can largely outperform state-of-the-art methods and produce more visually favorable results. The codes are rleased at \url{https://github.com/greatlog/RealDAN.git}.
}



\maketitle

\section{Introduction}\label{sec:introduction}

	Single image super-resolution (SR) aims to recover the high-resolution (HR) version of a given degraded low-resolution (LR) image. It has wide applications in video enhancement, medical imaging, as well as security and surveillance imaging. Generally, the degradation process can be formulated as
\begin{equation}
	\mathbf{y} = [(\mathbf{x}\otimes \mathbf{k})\downarrow_{s} + \mathbf{n} \label{downsample}]_q
\end{equation}
where $\mathbf{x}$ is the original HR image, $\mathbf{y}$ is the degraded LR image, $\otimes$ denotes the two-dimensional convolution of $\mathbf{x}$ with blur kernel $\mathbf{k}$, $\mathbf{n}$ denotes the random noise, $\downarrow_{s}$ denotes the standard $s$-fold downsampler (keeping only the upper-left pixel for each distinct $s\times s$ patch)~\cite{usr}, and $[\cdot]_q$ denotes JPEG compression with quality factor $q$. Then SR refers to the process of recovering $\mathbf{x}$ from $\mathbf{y}$. In a blind case, not only the HR image $\mathbf{x}$, but also the degradation parameters ($\mathbf{k}$, $\mathbf{n}$ and $q$) are unknown, which makes blind SR a quite challenging task~\cite{baker2002limits}.

Most previous methods try to decompose blind SR into two relatively easier steps: 1) \textit{degradation estimation}, and 2) restoring the SR image with the estimated degradation, which is called \textit{non-blind SR}. Following this framework, degradation estimation and non-blind SR have been independently studied for years, and many successful methods have been proposed in both research fields, such as Michaeli \etal~\cite{nonpara}, KernelGAN~\cite{kernel_gan} for degradation estimation and DPSR~\cite{dpsr}, USRNet~\cite{usr} for non-blind SR. 

However, two problems may exist in this framework: 1) The degradation-estimation model and the SR model are independently optimized. It is likely that the two models could not well cooperate, \ie small deviations of degradation estimation may lead to terrible SR results. 2) Degradation estimation is inherently ambiguous due to the information loss during downscaling~\cite{least_square}. It is extremely difficult to get an accurate degradation in the absence of original HR information. A straightforward idea to address the second problem is using SR results to help improve the accuracy of degradation estimation. However, as we have described in the first problem, a good SR result also requires an accurate degradation in the first place. As a result, previous methods often fall into a dilemma: how to simultaneously improve the accuracy of degradation estimation and SR performance?

To break this dilemma, instead of considering these two steps separately, we adopt an alternating optimization algorithm, which can estimate the degradation and restore the SR image in a single model. In detail, we design two convolutional neural modules, namely \textit{Restorer} and \textit{Estimator}. \textit{Restorer} restores the SR image based on the degradation estimated by \textit{Estimator}, and the restored SR image is further used to help \textit{Estimator} estimate a more accurate degradation. Once the degradation is manually initialized, the two modules can well corporate with each other to form a closed loop, which can be iterated over and over. We fix the number of iterations and unfold the iterating process to form an end-to-end trainable network, which we call the deep alternating network (DAN). To ensure the convergence of the iteration, DAN is directly supervised at the last iteration during training. Thus, both \textit{Restorer} and \textit{Estimator} may learn to substantially improve their results during the iterating and finally coverage to a stable point. In the framework of DAN, \textit{Estimator} can utilize the information of intermediate SR results, which makes the degradation estimation easier. More importantly, as \textit{Restorer} and \textit{Estimator} are jointly optimized, they may get more tolerant of the deviations of each other and cooperate well to achieve a better final result.

We need to note that a preliminary version of this work has been presented as a conference paper~\cite{dan}, which is denoted as DAN-Pre in this paper. In the current version, we incorporate additional content in significant ways.

Firstly, DAN-Pre can only process blurry LR images (with mild additive white Gaussian noise (AWGN)). While in DAN, we consider much more complex degradations, including multiple blur, resize, noise, and JPEG compression. To deal with such complex degradations, we parameterize the whole degrading pipeline and each random degradation can be represented by a unique vector that is computable for both \textit{Restorer} and \textit{Estimator}.

Secondly, in DAN-Pre, the \textit{Restorer} and \textit{Estimator} are iterated in the image space and degradation space respectively, while in DAN, the two modules are iterated in the feature space. In this way, richer information can be passed between different iterations, which may make the training process more stable and get the whole network better optimized. Moreover, iterating in the feature space also saves computations as we do not need to compute the output in each iteration. 

Thirdly, we reorganize the experiments and make more comprehensive comparisons with the most recent state-of-the-art methods. We also add more experiments to better analyze the proposed method.

We summarize our contributions into following points:
\begin{itemize}
	\item [1.] We propose an alternating optimization algorithm that considers the degradation estimation and SR in a single network, in which way, both modules can utilize the intermediate results of each other and could get well compatible to produce better final results than previous two-step solutions. 
	\item [2.] To the best of our knowledge, the preliminary version of this work~\cite{dan} proposes the first end-to-end network for blind SR, which largely simplifies and accelerates the training and inference of blind SR methods.
	\item [3.]  We parameterize the complex degradation process (including multiple blur, resize, noise, and compression) and make it computable for convolutional networks. To the best of our knowledge, the proposed DAN is the first network that can simultaneously estimate the complex degradations and restore SR images for real-world blind SR.
	\item [4.] We design two convolutional neural modules, which can be alternated repeatedly to estimate the degradation and restore the SR image respectively. 
	\item [5.] Extensive experiments on synthetic and real-world images show that our model can largely outperform state-of-the-art blind-SR methods and produce more visually favorable results.
\end{itemize}

\section{Related Work}
In the recent decade, deep-learning (DL)-based SR methods~\cite{rcan,esrgan,swinir,survey,zhang2022memory,zhou2022memory} have made remarkable achievements and have shown great advantages against traditional methods~\cite{sc,kk,a+}. Thus, we mainly discuss DL-based SR methods in this paper.

\subsection{SR for Bicubically Downscaled Images}
DL-based SR methods usually require a large number of paired HR-LR images as training samples. However, these paired samples are difficult to be collected in the real world. Consequently, synthetic data is usually used as an alternative. Early researchers consider only the simplest case, \ie the LR images are obtained by downscaling the HR images with bicubic interpolation~\cite{srcnn}. In this way, a large number of training samples can be cheaply synthesized. In this case, as data is not the concern, most researchers concentrate on designing the structures of SR networks. In SRCNN~\cite{srcnn}, Dong \etal propose the first convolutional neural network (CNN) for SR, which has only three convolutional layers. In the following years, many CNN-based SR methods~\cite{dbpn,meta_sr,imdn,carn,idn} have been proposed and strategies such as  post-upsampling~\cite{fsrcnn}, residual learning~\cite{vdsr}, and pixel-shuffle operation~\cite{pixel_shuffle} nearly become the default choices for building an SR network. After the proposal of RCAN~\cite{rcan}, RRDB~\cite{esrgan} and SAN~\cite{san}, the CNN-based SR performance even starts to get saturate on common benchmark datasets. Recently, some transformer~\cite{transformer}-based methods, such as IPT~\cite{ipt} and SwinIR~\cite{swinir} further advance the performance for bicubically downscaled images.

However, although these methods perform well for super-resolving bicubically downscaled images, it is still difficult for them to get applied in real scenarios. The degradations in real scenarios are various and unknown, which are much more complex than the bicubic downscaling. Consequently, due to the domain gap between real and synthesized data, methods designed for bicubically downscaled images will suffer serve performance drop in real applications~\cite{bridge,cycleSR}.  To address this issue, researchers begin to work on more challenging cases where degradations of test images are unknown, \ie blind SR.

\subsection{SR for Blurry Downscaled LR Images}\label{sec:blurry_sr}
Early blind SR methods walk a step further in practicability: they also consider cases where LR images are blurred by various kernels (sometimes AWGN is also considered) before downscaling. Compared with SR for bicubically downscaled images, apart from the HR image $\mathbf{x}$, there is one more unknown variable, i\ie the blur kernel $\mathbf{k}$. Thus, this problem usually involves two steps: kernel estimation and SR with the given kernel. Some methods focus on only one of them and some try to solve them simultaneously.

\vspace{0.02\linewidth}\noindent\textbf{Kernel Estimation.}
Some methods only focus on kernel estimation. As this is an ill-posed problem~\cite{levin,levin2011efficient}, some priors are usually needed to get it properly solved. In~\cite{nonpara}, a non-parametric method is used by utilizing the patch recurrence between the test image and its downscaled version. A similar idea is also adopted in~\cite{kernel_gan}, but powered with neural networks and adversarial training~\cite{gan}. Another widely used prior is the extreme channel priors. In~\cite{pan,pan2016blind},  Pan \etal firstly propose the dark channel prior, \ie the dark channel in a natural image is usually sparse, which can be used for solving the blur kernel from a blurred image. In~\cite{extreme,cai2020dark}, the bright channel prior is further proposed and the idea is augmented to extreme channel priors. 

\vspace{0.02\linewidth}\noindent\textbf{SR with given kernels.}
Some methods assume that the kernel estimation has been solved and only focus on SR with given kernels. In~\cite{zssr_pre,zssr,mzsr}, the blur kernel is used to down sample images and synthesize training samples, which can be used to train a specific model for a given kernel and LR image. In SRMD~\cite{srmd}, the kernel and LR image are directly concatenated at the first layer of a CNN. Thus, the SR result can be closely correlated to both LR image and blur kernel. In DPSR~\cite{dpsr}, Zhang \etal propose a method based on the ADMM algorithm. They interpret this problem as MAP optimization and solve the data term and prior term alternately. A similar idea is adopted in USRNet~\cite{usr}. These methods can achieve remarkable performance as long as the ground-truth blur kernel is known. However, in real applications, the blur kernels are predicted by kernel-estimation methods and are likely to be biased from the ground-truth ones. This bias will cause a serious performance drop when the kernel-estimation step and the SR step are combined.

\vspace{0.02\linewidth}\noindent\textbf{Integrated SR with Kernel Estimation.}
To avoid the performance drop when the two steps are combined, recent methods try to integrate them into a single model, which consists of a kernel-estimation module and a SR module. In~\cite{ikc}, Gu \etal propose to finetune the kernel-estimation module together with the SR module to get them more compatible. Its improvement is evident, but the training and inference are complicated and time-consuming. In the preliminary version of this paper~\cite{dan}, we adopt an alternating optimization algorithm and propose the first end-to-end method for integrated SR with kernel estimation, which further improves the compatibility of the two modules. And since then the end-to-end framework becomes the prevalent choice for blind SR methods~\cite{koalanet,dipfkp}. In DASR~\cite{dasr}, the kernel representations are firstly extracted via contrastive learning and then input to a SR module. In DCLS~\cite{least_square}, the kernel is firstly estimated by the dynamic deep linear module and then input into a deep-constrained-least-squares module to help restore the SR image. However, the kernel-estimation module in these methods utilizes only information from the LR images, which may limit the accuracy of kernel estimation and the final SR performance. While in DAN, the accuracy of degradation estimation may be improved with the help of intermediate SR results.

\subsection{SR for Real Images}
Although the blurry LR image is a better assumption than the bicubically downscaled one, it may still be far from real images ~\cite{real-esrgan,bsrgan,inter_clr}. The domain gap between training and testing data will largely destroy the practicability of  SR methods. To address this issue, some researchers try to manually collected paired HR-LR samples~\cite{cdc,lp-kpn}. However, it is expensive and time-consuming, and there may still have a domain gap between images collected by different cameras. Recently, in BSRGAN~\cite{bsrgan} and Real-ESRGAN~\cite{real-esrgan}, researchers propose to synthesize training samples by much more complex degradations, including multiple blur, resize, noise, and compression. In this way, the synthesized data may be diverse enough to include most cases in real scenarios. And the SR model trained with these samples may be practical enough in applications. However, the two methods are degradation-unaware, \ie directly super-resolving the LR image regardless of its degradation. Consequently, they may fail to exploit a more general relationship between SR under various degradations and could only achieve sub-optimal results. While in the current version of DAN, we re-parameterize the complex degradation process and adopt an alternating optimization algorithm to simultaneously do SR and degradation estimation, which may lead the SR module to be more degradation-specific and achieve better performance.

\begin{figure*}
	\centering
	\includegraphics[width=\linewidth]{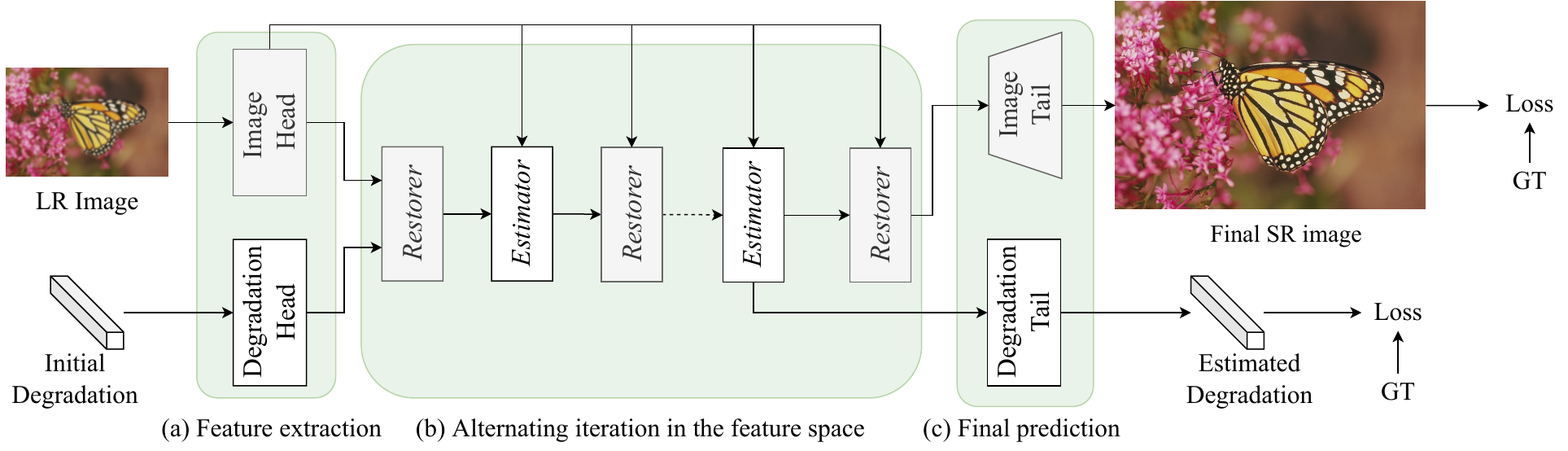}
	\caption{The overview of the deep alternating network (DAN).}\label{overview}
\end{figure*}

\section{End-to-End Alternating Optimization}
\subsection{Formulation}\label{formulation}
Generally, the degradation process shown in Equation~\ref{downsample} can be formulated as: 
\begin{equation}
	\mathbf{y} = \mathcal{D}(\mathbf{x}; \mathbf{\theta}),
\end{equation}
where $\mathcal{D}$ is the degradation function, and $\mathbf{\theta}$ is the involved parameter. Then the blind SR problem can be mathematically expressed an optimization problem in the Maximum A Posterior (MAP) framework~\cite{zhang2017learning}:
\begin{equation}
	\mathbf{x}, \mathbf{\theta}= \mathop{\arg\min}_{\mathbf{x}, \mathbf{\theta}}  \|\mathbf{y} -  \mathcal{D}(\mathbf{x}; \mathbf{\theta}) \|_1.
\end{equation}
However, as there are too many unknown variables, this optimization problem is still ill-posed and has an infinite number of solutions~\cite{baker2002limits}. To get it properly solved, some prior terms are usually added~\cite{selfdeblur,deblurpair}:
\begin{equation}
	\mathbf{x}, \mathbf{\theta}= \mathop{\arg\min}_{\mathbf{x}, \mathbf{\theta}}  \|\mathbf{y} -  \mathcal{D}(\mathbf{x}; \mathbf{\theta}) \|_1 + \phi(\mathbf{x}) + \psi(\mathbf{\theta}), 
\end{equation}\label{eq:ori_problem}
where $\phi(\mathbf{x})$ denotes the prior for HR image, and $\psi(\mathbf{\theta})$ represents the prior for degradation parameter. 

\subsection{Two-step Solution}
In the context of SR for blurry downscaled images, $\mathcal{D}$ represents blur and downscaling. And $\mathbf{\theta}$ represents the blur kernel $\mathbf{k}$ and downscaling factor $s$. In many methods, $\mathbf{k}$ is assumed to be a Gaussian kernel~\cite{ikc,dasr}. Such prior makes it easier to solve the degradation parameter, which can be further used to solve the HR image. In this case, previous usually adopt a two-step solution:
\begin{equation}
	\left \{
	\begin{aligned}
		\mathbf{\theta} &= \mathcal{K}(\mathbf{y})\\
		\mathbf{x} &=  \mathop{\arg\min}_{\mathbf{x}}  \|\mathbf{y} -  \mathcal{D}(\mathbf{x}; \mathbf{\theta}) \|_1 + \phi(\mathbf{x}) 
	\end{aligned}
	\right.
\end{equation}
where $\mathcal{K}(\cdot)$ denotes the function that estimates $\mathbf{\theta}$, \ie blur kernel $\mathbf{k}$ in this case, from $\mathbf{y}$. And the second step is usually solved by a non-blind SR method such as DPSR~\cite{dpsr}, USRNet~\cite{usr}, \etc. 

As we have mentioned in Sec~\ref{sec:blurry_sr}, the two steps are independent research fields in most cases. Both of them only consider the performance under their own given conditions, while ignoring the overall performance. This two-step solution has its drawbacks threefold. Firstly, this algorithm usually requires the training of two or even more models, which is rather complicated. Secondly, $\mathcal{K}(\cdot)$  can only utilize information from $\mathbf{y}$. However, this is also an ill-posed problem:  $\mathbf{\theta}$ could not be properly solved without information from $\mathbf{x}$. At last, the non-blind SR model for the second step is trained with ground-truth degradations. While during testing, it only has access to degradations estimated in the first step. The difference between ground-truth and estimated degradations will usually cause a serve performance drop of the non-blind SR model~\cite{ikc}. Moreover, in cases of more complex degradations, the deviations of estimated degradations are likely to be larger, which may further destroy the SR performance of the second step.

\subsection{Unfolding the Alternating Optimization}\label{sec:unfold}
Towards the drawbacks of two-step solution, we propose an end-to-end network that can largely alleviate these issues. Specifically, we still split it into two subproblems. But instead of solving them in sequential, we adopt an alternating optimization algorithm, which restores the SR image and estimates the degradation alternately. The mathematical expression is
\begin{equation}
	\left \{
	\begin{aligned}
		\mathbf{\theta}_{i+1} &=  \mathop{\arg\min}_{\mathbf{\theta}}  \|\mathbf{y} -  \mathcal{D}(\mathbf{x}_{i}; \mathbf{\theta}) \|_1  + \psi(\mathbf{\theta}) \\
		\mathbf{x}_{i+1} &= \mathop{\arg\min}_{\mathbf{x}}  \|\mathbf{y} -  \mathcal{D}(\mathbf{x}; \mathbf{\theta}_{i}) \|_1 + \phi(\mathbf{x}).
	\end{aligned}
	\right.
\end{equation}
We define two solvers, namely \textit{Estimator} and \textit{Restorer} for the two subproblems respectively. And The blind SR problem may be solved by repeatedly iterating the two solvers. Since the prior terms, \ie $\psi(\mathbf{\theta})$ and $\phi(\mathbf{x})$, are difficult to be mathematically expressed (unless in some simple cases), it is also hard to find the analytic solutions for both solvers. Given the remarkable achievements of DL networks, we try to construct both solvers with convolutional neural modules. We hope that once the two modules are trained, they could automatically solve the sub-problems respectively.

As shown in Fig~\ref{overview} (a), we firstly initialize the degradation as $\mathbf{\theta}_0$ (set as learnable and initialized as null-vector in our experiments) and then encode the LR image and initial degradation to the feature space with two head modules: 
\begin{equation}
	\mathbf{f}^{x}_0 = H^x(\mathbf{y}), \quad \mathbf{f}^{\theta}_0 = H^{\theta}(\mathbf{\theta}_0),
\end{equation}
where $H^x$ and $H^{\theta}$ are the head modules for images and degradations respectively, $\mathbf{f}^x_0$ and $\mathbf{f}^{\theta}_0$ are the initial features of images and degradations respectively. As shown in Fig~\ref{overview} (b), we then iterate the \textit{Restorer} $R$ and the \textit{Estimator} $E$ in the feature space to alternatively solve the features for images and degradations:
\begin{equation}
	\left\{
	\begin{aligned}
		&\mathbf{f}^{x}_{i+1} = R(\mathbf{y}, \mathbf{f}^{\theta}_{i}) \\
		&\mathbf{f}^{\theta}_{i+1} = E(\mathbf{y}, \mathbf{f}^{x}_{i}).
	\end{aligned}
	\right.
\end{equation}
After $T$ iterations, the image features are super-resolved to the HR image, and the degradation features are regressed to the estimated degradation (shown in Fig~\ref{overview} (c)):
\begin{equation}
	\mathbf{x} = T^x(\mathbf{f}^{x}_{T}), \quad \mathbf{\theta} = T^{\theta}(\mathbf{f}^{\theta}_{T}),
\end{equation}
where $T^x$ and $T^{\theta}$ are the tail modules for the images and degradations respectively. We fix the number of iterations and unfold the iterating process. Then the whole pipeline can form an end-to-end trainable network, which is called the deep alternating network (DAN). Since we hope that the final iteration results could converge to a stable point, we supervise DAN at the end by the ground-truth degradation and HR image. 

\subsection{Discussion}
Compared with previous two-step methods, DAN has its benefits threefold: 1) DAN can be trained and tested in an end-to-end manner, which largely simplifies and accelerate blind SR methods; 2) the \textit{Estimator} can utilize the information from intermediate SR results, which makes the degradation estimation easier; 3) \textit{Estimator} and \textit{Restorer} are jointly optimized, and thus they may get more compatible with each other and cooperate better to achieve preferable final results. 

It should be noted that despite both DAN and IKC utilizing an iterative approach, DAN and IKC both adopt an iterative strategy, their optimization algorithm differs significantly. In IKC, the SR module is initially trained as a non-blind SR model and then kept fixed during the training of the kernel correction module. This sequential training process in IKC can be cumbersome. Additionally, the SR module in IKC only has access to ground-truth degradations during its training phase, rather than estimated degradations. Consequently, the compatibility between the SR module and kernel-correction module may be compromised, leading to sub-optimal performance. In contrast, in DAN, the \textit{Restorer} and \textit{Estimator} are jointly optimized in an unfolded network. Thus both modules are trained in an end-to-end manner, simplifying the training process. Moreover, the SR module (\ie the \textit{Restorer}) in DAN is trained with estimated degradations, which may lead to better compatibility with the \textit{Estimator} and potentially achieve superior results.

After the proposal of the preliminary version of this work, many other end-to-end blind SR methods~\cite{dasr,least_square} are also proposed. However, their degradation-estimation modules usually utilize only the limited information from the LR image, which may restrict their performance.

Compared with the preliminary version of this work, the current framework has two main differences. Firstly, the initial degradation is manually set in the preliminary version while it is learnable in the current version. In this way, it is more likely that DAN could adaptively find a good initial point for the alternating optimization, which may help improve the final results. Secondly, in the preliminary version, the information of different iterations can only be passed through the SR image (degradation). While in the current version, \textit{Estimator} and \textit{Restorer} are iterated in the feature space and predict the final results only at the last iteration. In this way, richer information could be passed through intermediate features. And this strategy could also reduce the redundant computations of final results.
\begin{figure}[t]
	\centering
	\includegraphics[width=\linewidth]{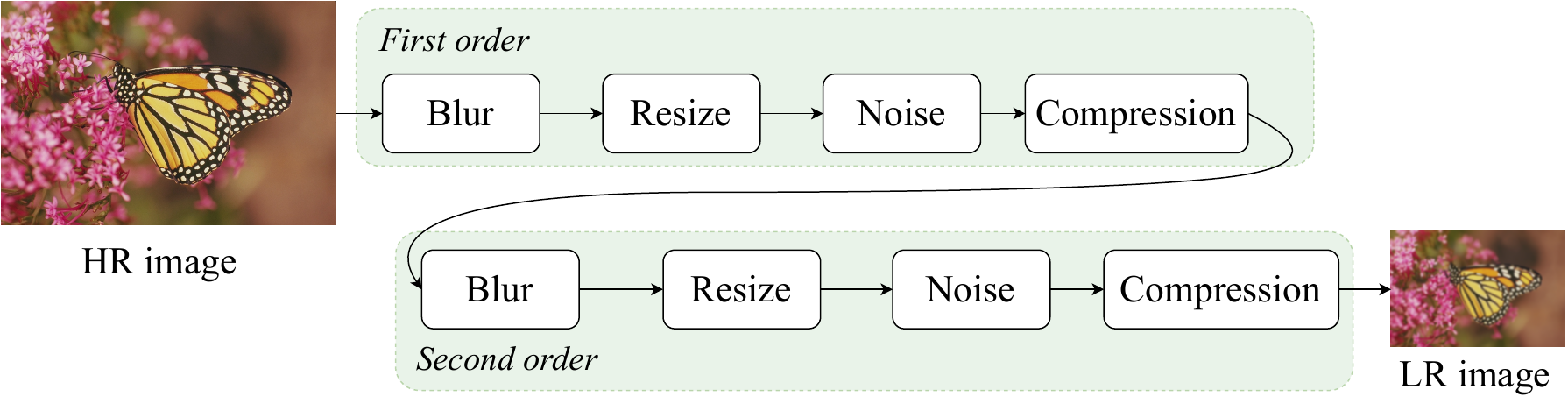}
	\caption{The multi-order degradation model.}\label{fig:deg_pipeline}
\end{figure}
\begin{figure*}[t]
	\centering
	\includegraphics[width=\linewidth]{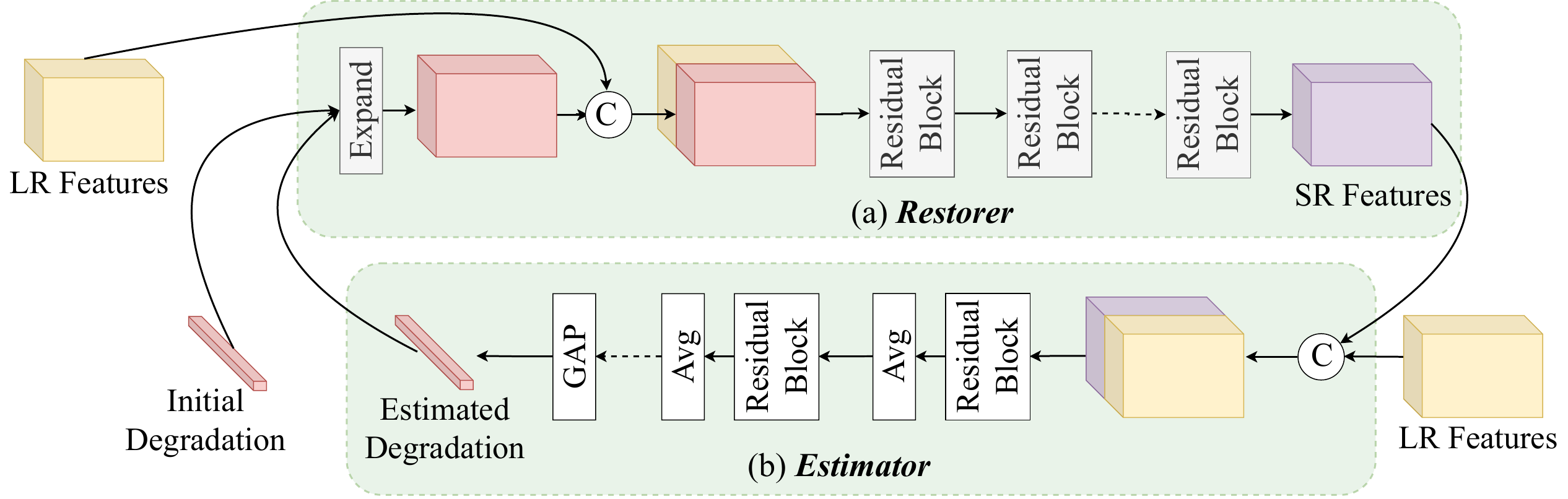}
	\caption{The frameworks of  (a) \textit{Restorer} and (b) \textit{Estimator}. $\copyright$  denotes concatenating in the channel dimension. And 'Expand' means expanding the degradation features in spatial dimensions. `Avg' denotes average pooling, and `GAP' denotes global average pooling.} \label{modules}
\end{figure*}

\subsection{Reparamizaton of the Degradation}\label{sec:rep_deg}
Previous methods~\cite{ikc,dan} consider only blurry LR images, in which case the degradation is parameterized as the blur kernel. While in real applications, degradations are much more complex and are difficult to be parameterized and calculated. In Real-ESRGAN~\cite{real-esrgan}, Wang \etal propose a multi-order degradation model(shown in Fig~\ref{fig:deg_pipeline}), which includes multiple blur, resize, noise, and compression. Although each single degradation operation is simple, they could be integrated to well simulate the complex real degradations. We adopt the same degradation model in work and we try to re-parameterize this model by parameterizing each degradation operation.

\subsubsection{Blur}
In the multi-order degradation model, the original HR image is firstly blurred. In~\cite{real-esrgan}, three kinds of  blur kernels are considered, including generalized Gaussian kernels~\cite{generalized_gauss}, generalized plateau kernels, and what they call $sinc$ kernels. The generalized kernel can be mathematically expressed as
\begin{equation}
	\mathbf{k}(i, j) = \exp(-\frac{1}{2}(\begin{bmatrix} i & j\end{bmatrix}\mathbf{\Sigma}^{-1}\begin{bmatrix} i\\j \end{bmatrix})^{\beta}), \label{eq:gaussian_kernel}
\end{equation}
where ${i, j}$ is coordinate of $\mathbf{k}$, $\mathbf{\Sigma}$ is the covariance matrix, and $\beta$ is the shape parameter. If $\beta=1$, $\mathbf{k}$ becomes a common Gaussian kernel. Similarly, the plateau kernel can be mathematically expressed as
\begin{equation}
	\mathbf{k}(i, j) = \frac{1}{1 + (\begin{bmatrix} i & j\end{bmatrix}\mathbf{\Sigma}^{-1}\begin{bmatrix} i\\j \end{bmatrix})^{\beta}}.
\end{equation}
For both Gaussian and plateau kernels, the covariance matrix can be expressed as:
\begin{equation}
	\mathbf{\Sigma} = \begin{bmatrix}\sigma_x & 0 \\ 0 & \sigma_y \end{bmatrix} 
	\begin{bmatrix}\cos\theta & -\sin\theta \\ \sin\theta & \cos\theta\end{bmatrix}.
\end{equation}
Thus, either the Gaussian kernel or plateau kernel can be parameterized as four parameters, \ie $\sigma_x$, $\sigma_y$, $\theta$, and $\beta$.  The $sinc$ kernel is expressed as
\begin{equation}
	\mathbf{k}(i, j) = \frac{\omega_c}{2\pi\sqrt{i^2 + j^2}}J_1(\omega_c\sqrt{i^2+j^2}),
\end{equation}
where $\omega_c$ is the cutoff frequency, and $J_1$ is the first order Bessel function of the first kind. Thus, the $sinc$ kernel can be parameterized as one parameter $\omega_c$.

We further define a variable $k_s$ to indicate the kernel size and a vector $[k_g, k_c]$ to indicate the type of the blur kernel. Specifically, $k_g=0, k_c=0$ indicates a plateau kernel, $k_g=1, k_c=0$ indicates a Gaussian kernel, and $k_g=0, k_c=1$ indicates a $sinc$ kernel. Then, the blur kernel can be parameterized as
\begin{equation}
	\mathbf{B} = \begin{bmatrix}k_g, k_c, k_s, \sigma_x, \sigma_y, \theta, \beta, \omega_c\end{bmatrix}.
\end{equation}
We need to note when the $\mathbf{b}$ represents a $sinc$ kernel, we set $\sigma_x=\sigma_y=\theta=\beta=0$, since a $sinc$ kernel dose not have those parameters.

\subsubsection{Resize}
After being blurred, the image is  then resized. In~\cite{real-esrgan}, there are three different modes of resizing, namely \textit{area}, \textit{bilinear}, and \textit{bicubic}. Each image is resized via the mode randomly chosen from them. We use a one-hot vector $[r_{area}, r_{bil}, r_{bic}] $ to indicate the resizing mode. In detail, $[1, 0, 0]$ indicates the \text{area} mode, $[0, 1, 0]$ indicates the \textit{bilinear} mode, and $[0, 0, 1]$ indicates the \textit{bicubic} mode. We further denote the scale factor as $s$. And the resizing operation can be parameterized as:
\begin{equation}
	\mathbf{R} = \begin{bmatrix}r_{area}, r_{bil}, r_{bic}, s\end{bmatrix}.
\end{equation}

\subsubsection{Noise}
In both Real-ESRGAN and BSRGAN, two kinds of noises are considered, \ie the Gaussian noise and the Poisson noise. The Gaussian noise $\mathbf{n}_g \sim \mathcal{N}(0, \sigma_g)$ is parameterized by the standard deviation $\sigma_g$. And the Poisson noise $\mathbf{n}_p$ is parameterized by the noise level $\lambda$~\cite{real-esrgan,unprocessing_raw}. We define a variable $n_t$ to indicate the type of the noise. Specifically, $n_t=1$ indicates Gaussian noise, and $n_t=0$ indicates Poisson noise. The are also two kinds of color for the noise, \ie gray and RGB. Thus, we further define a variable $n_c$ to indicate the noise color ($1$ for RGB and $0$ for gray). Finally, the noise $\mathbf{n}$ can be parameterized as
\begin{equation}
	\mathbf{N} = \begin{bmatrix}n_t, n_c, \sigma_g, \lambda\end{bmatrix}.
\end{equation}
We need to note when $n$ represents the Gaussian noise, we have $\lambda=0$  since there is no Poisson noise. Similarly, when $n$ represents the Poisson noise, we also have $\sigma_g=0$.

\subsubsection{JPEG Compression}
In real applications, some LR images will be compressed by the JPEG compression~\cite{jpeg}. We define a variable $j$ to indicate whether the image is compressed ($1$ for yes and $0$ for no). For the compressed images, the JPEG compression can be parameterized by the quality factor $q$ (sampled in $[1, 100)$). Thus, the compression operation $\mathbf{J}$ can be parameterized as
\begin{equation}
	\mathbf{J} = \begin{bmatrix}j, q\end{bmatrix}.
\end{equation}
We need to note when $j=0$, we have $q=100$, which indicates an uncompressed image.

\subsubsection{Multi-order Degradation}
As shown in Fig~\ref{fig:deg_pipeline}, in the multi-order degradation model, the HR image will go through the \textit{blur-resize-noise-compression} pipeline for twice before it is degraded to a LR image. As we have discussed above, this pipeline can be parameterized as
\begin{equation}
	\mathbf{D} = \begin{bmatrix} \mathbf{B}, \mathbf{R}, \mathbf{N}, \mathbf{J}\end{bmatrix}, 
\end{equation}
which is a 1-dimension vector. Thus, the whole degradation model can be parameterized as
\begin{equation}
	\mathbf{\theta} = \begin{bmatrix} \mathbf{D}_1,  \mathbf{D}_2\end{bmatrix},\label{eq:deg_param}
\end{equation}
where $\mathbf{D}_1$ and $\mathbf{D}_2$ represent the parameters of the first and the second order degradation respectively, and $\mathbf{\theta}$ is also a 1-dimension vector. In this way, each degradation can be uniquely represented and calculated.

\subsection{Instantiation}\label{sec:instantiation}
\textbf{Head modules,}
As we have described in Sec~\ref{sec:unfold} and Fig~\ref{overview}, the features of the LR image and degradation are firstly extracted by two \textit{head} modules respectively. The image head in Fig~\ref{overview} is constructed by one convolutional layer. While the degradation head is constructed by one fully connected layer since the degradation is a one-dimension vector. The extracted shallow features are then input to the \textit{Restorer} and \textit{Estimator}.

\vspace{0.02\linewidth}\noindent
\textbf{\textit{Restorer} and \textit{Estimator.}}
Both of the \textit{Restorer} and \textit{Estimator} need to cope with two inputs. In the preliminary version~\cite{dan}, we propose a conditional residual block to help the two modules exploit the inter information of their two inputs and ensure that the output of the \textit{Restorer} (\textit{Estimator}) is closely related to both of its inputs. However, we later experimentally find that the same effect can be achieved by simple concatenation at the beginning of \textit{Restorer} (\textit{Estimator}).

For the \textit{Restorer}, as shown in Fig~\ref{modules} (a), we simply concatenate the LR feature and the degradation feature at the beginning and then restore them to the SR features via the body of the \textit{Restorer}. We need to note that the degradation feature is a one-dimension vector (Eq~\ref{eq:deg_param}). Thus we need to expand it spatially before it is concatenated with the LR feature. In this way, the body of \textit{Restorer} can take the benefits of the architectures in recent state-of-the-art SR models, such as EDSR~\cite{edsr}, RCAN~\cite{rcan}, and RRDB~\cite{esrgan}. For simplicity, we use $16$ residual blocks proposed in EDSR to construct the body of \textit{Restorer}.

For the \textit{Estimator}, as shown in Fig~\ref{modules} (b), the two inputs, \ie the LR features and the SR features, are also concatenated at the beginning. The body of \textit{Estimator} is constructed by residual blocks and average pooling. Since the degradation estimation usually requires global information, especially for estimation of noise level and quality factor of jpeg compression. Thus each residual block is followed by an average pooling (AvgP) layer to enlarge the receptive fields of  \textit{Estimator}. And at the end of  \textit{Estimator}, an global average pooling (GAP) layer is used to predict the degradation features. 

Compared with the DAN-Pre, the architectures of \textit{Restorer} and \textit{Estimator} are largely simplified, which further accelerates the training and testing of DAN (which will be discussed in Sec~\ref{sec:compare_speed}). While experiments in Sec~\ref{exp:compare_div2krk} show that the DAN can achieve even better results than DAN-Pre.

\vspace{0.02\linewidth}\noindent
\textbf{Tail modules.}
As shown in Fig~\ref{overview}, the SR feature and the degradation feature are input into two \textit{tail} modules. The image tail upscales the SR feature and reconstructs it to the SR image. While the degradation tail projects the degradation feature to the final estimated degradation. The image tail consists of a PixelShuffle~\cite{pixel_shuffle} layer and server convolutional layers. And the degradation tail consists of only fully connected layers.

\begin{table*}[t]
	\centering
	\caption{Quantitative comparison with state-of-the-art methods for blurry LR images. Average PSNR and SSIM results on DIV2KRK~\cite{kernel_gan} are reported. $\times2$ and $\times4$ denote the scale factors. $\uparrow$ denotes the higher the better. The best two results are indicated in bold and underlined respectively.} \label{tab:compare_div2krk}
	\setlength{\tabcolsep}{18pt}
	\resizebox{\linewidth}{!}{
		\begin{tabular}{lccccc}
			\toprule
			\multirow{2}{*}{Methods}
			& \multicolumn{2}{c}{$\times2$} &
			& \multicolumn{2}{c}{$\times4$} \\
			\cmidrule{2-3} \cmidrule{5-6} 
			& PSNR$\uparrow$& SSIM$\uparrow$&
			& PSNR$\uparrow$& SSIM$\uparrow$      \\
			\midrule
			\multicolumn{6}{l}{\textit{Two-step methods}}\\
			\midrule
			KernelGAN~\cite{kernel_gan}+ZSSR~\cite{zssr}
			& $30.36$   & $0.8669$   &  & $26.81$   & $0.7316$   \\
			KernelGAN~\cite{kernel_gan}+SRMD~\cite{srmd}
			& $29.57$   & $0.8564$   &  & $27.51$   & $0.7265$   \\
			KernelGAN~\cite{kernel_gan}+USRNet~\cite{usr}
			& -       & -          &  & $20.06$   & $0.5359$   \\
			 Michaeli~\cite{nonpara}+SRMD~\cite{srmd}
			& $25.51$   & $0.8083$   &  & $23.34$   & $0.6530$   \\
			Michaeli~\cite{nonpara}+ZSSR~\cite{zssr}
			& $29.37$   & $0.8370$   &  & $26.09$   & $0.7138$   \\ 
			\midrule
			\multicolumn{6}{l}{\textit{End-to-end methods}} \\
			\midrule
			Bicubic                  & $28.73$   & $08040$    &  & $25.33$   & $0.6795$   \\
			EDSR~\cite{edsr}
			&$32.42$&$0.9034$&  & $28.68$&$0.7883$            \\
			IKC~\cite{ikc}
			& $31.20$   & $0.8767$   &  & $27.69$   & $0.7657$   \\
			AdaTarget~\cite{adatarget}
			& -               & -               &  & $28.42$   & $0.7854$   \\
			KOALAnet~\cite{koalanet}
			& $31.89$   & $0.8852$   &  & $27.77$   & $0.7637$   \\
			DCLS ~\cite{least_square}
			& $\underline{32.75}$   & $\underline{0.9094}$   &  & $\mathbf{28.99}$ & $\underline{0.7946}$   \\
			DAN-Pre~\cite{dan}
			& $32.56$   & $0.8997$   &  & $27.55$   & $0.7582$   \\
			DAN                    
			&$ \mathbf{32.96}$  & $\mathbf{0.9114}$&   &$\underline{28.90}$ &$\mathbf{0.7961}$           \\
			\bottomrule
	\end{tabular}}
\end{table*}
\begin{figure*}[t]
	\centering
	\includegraphics[width=\linewidth]{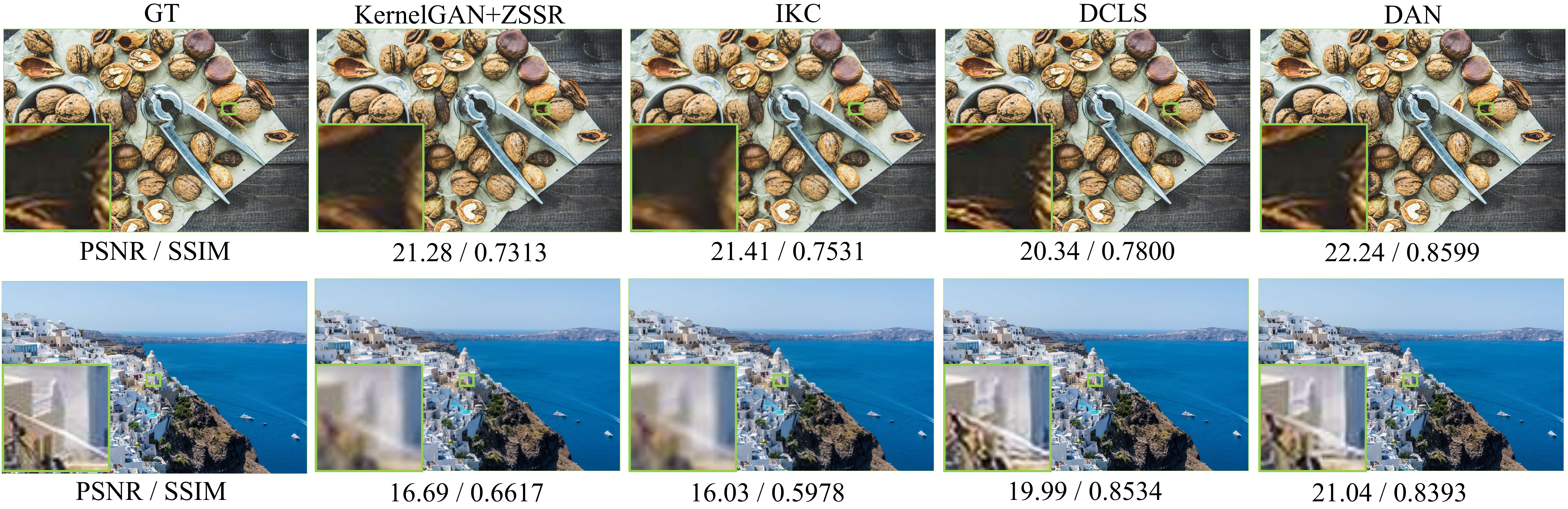}
	\caption{Visual comparisons between different methods on DIV2KRK for scale factor $\times2$. Best viewed in color.}\label{fig:vis_setting1_x2}
\end{figure*}
\begin{figure*}[t]
	\centering
	\includegraphics[width=\linewidth]{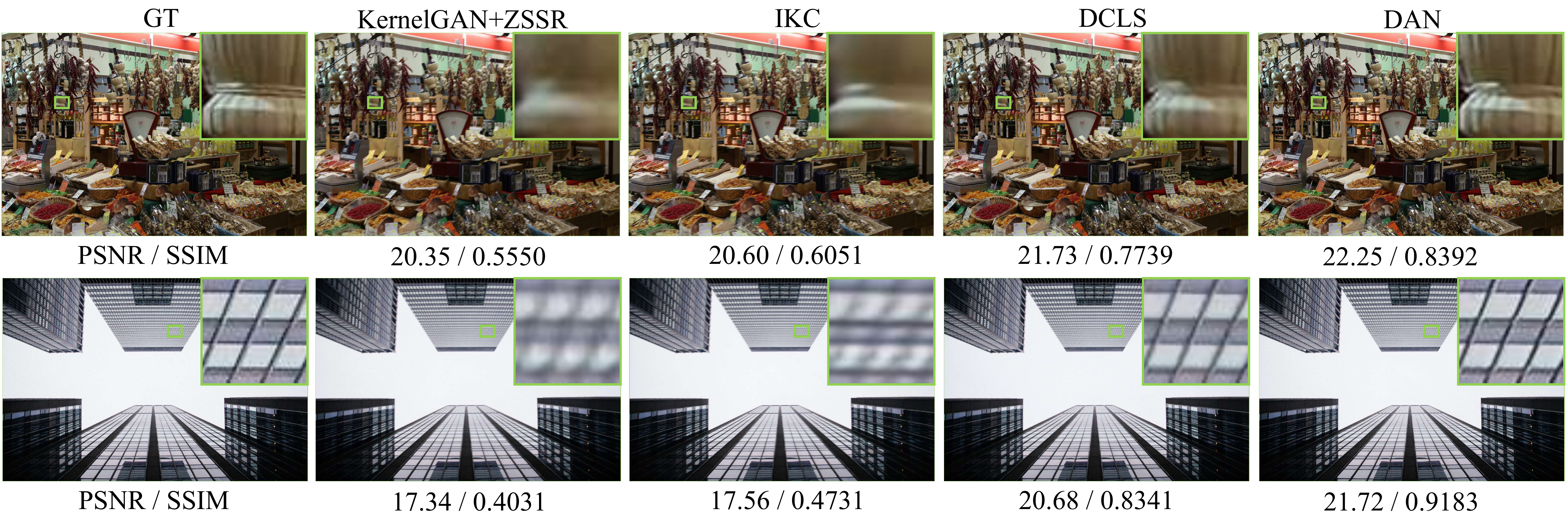}
	\caption{Visual comparisons between different methods on DIV2KRK for scale factor $\times4$. Best viewed in color.}\label{fig:vis_setting1_x4}
\end{figure*}

\section {Experiments}
\subsection{Experimental Setup}
\textbf{Datasets}
Following previous blind SR methods, DAN is also trained on datasets synthesized from the HR images in the training set of DIV2K~\cite{div2k} and Flickr2K~\cite{flickr2k}. The two datasets contain $3450$ HR images ($800$ from DIV2K and $2650$ from Flickr2K). Based on the discussion in Sec~\ref{sec:rep_deg}, DAN can deal with LR images with various degradations. However, most previous SR methods can only handle blurry LR images. Thus, to make comprehensive comparisons with other methods, we train DAN with two different settings.

Under the first setting, DAN is trained and evaluated on blurry LR images. Following the setting in~\cite{kernel_gan,dasr,least_square}, The training dataset is synthesized by various random anisotropic Gaussian kernels. For scale factor $\times2$ and $\times4$, the kernel size set as $11\times11$ and $31\times31$ respectively. The $\beta$ in Eq~\ref{eq:gaussian_kernel} is set as $1$, $\sigma_x$ and $\sigma_y$ are uniformly sampled in $(0.6, 5.0)$, and $\theta$ is uniformly sampled in $[-\pi, \pi]$. We also apply uniform multiplicative noise (up to $25\%$ of each pixel value of the kernel) and normalize it to sum to one. Under this setting, DAN is evaluated on the DIV2KRK~\cite{kernel_gan} dataset.

Under the second setting, DAN is trained on the dataset synthesized by more complex degradations. We use the same degradation model in Real-ESRGAN~\cite{real-esrgan} to synthesize HR-LR training pairs. The degradation is parameterized as we described in Sec~\ref{sec:rep_deg}. In this case, DAN is evaluated on real images from the track2 of NTIRE2020~\cite{ntire2020} (denoted as 2020Track2), and the RealSRSet collected in~\cite{bsrgan}. The 2020Track2 contains $100$ real images taken by iPhone, and RealSRSet contains $20$ real images either downloaded from the internet or directly chosen from existing testing datasets~\cite{dped,martin2001database,manga109,ffdnet}. However, those real images do not have corresponding ground-truth HR images, which makes it difficult to quantitatively measure the performance of different methods. Thus, for more comprehensive comparisons, we also synthesize a validation dataset, namely DIV2K-Real, with the same degradation model in~\cite{real-esrgan}.

\vspace{0.02\linewidth}\noindent
\textbf{Evaluation metrics.}
For validation sets that have ground-truth HR images, we use PSNR, SSIM~\cite{ssim}, and LPIPS~\cite{lpips}(the network is set as AlexNet~\cite{alex}) to evaluate the performance of different methods. We need to note that both PSNR and SSIM are calculated on the Y channel (\ie luminance) of transformed YCbCr space. For those real images that have no ground-truth references,  the non-reference metrics, \ie NIQE~\cite{niqe}, NRQM~\cite{nrqm}, and PI~\cite{pi} are used.

\vspace{0.02\linewidth}\noindent
\textbf{Loss functions.}
As shown in Fig~\ref{overview}, DAN is supervised by the ground-truth HR image and degradation. The supervision of HR image is applied via L1 loss, while the supervision of degradation is applied via L2 loss. In this paper, we also study the GAN~\cite{gan,srgan} version of DAN, which is denoted as DAN-GAN. Following the setting in RealESRGAN~\cite{real-esrgan}, DAN-GAN is also supervised by additional perceptual loss and adversarial loss. The perceptual loss is calculated in the feature space of VGG19~\cite{vgg}. The discriminator that is used to calculate the adversarial loss adopts the same U-Shape architecture in~\cite{real-esrgan}.

\vspace{0.02\linewidth}\noindent
\textbf{Training details.}
During training, the LR images are cropped into patches of $48\times48$. For scale factor $\times4$, the  HR images are cropped into $192\times192$, and for scale factor $\times2$, the HR images are cropped into $96\times96$. The batch size is set as $64$. The model is trained for $6\times10^5$ steps. The learning rate is initialized as $2\times10^{-4}$ and is decayed by half every $2\times10^5$ steps. We use Adam~\cite{adam} as the optimizer. And all models are trained on $4$ RTX 3090 GPUs.

\begin{table*}[t]
	\centering
	\caption{Quantitative comparison with state-of-the-art methods for real-world LR images. Average results for $\times4$ models are reported on DIV2K-Real,  the track2 of NTIRE2020, and RealSet~\cite{bsrgan}. $\uparrow$ denotes the higher the better, and $\downarrow$ denotes the lower the better. The best two results are indicated in bold.} \label{tab:compare_real}
	\setlength{\tabcolsep}{0.2cm}
	\resizebox{\linewidth}{!}{
		\begin{tabular}{lccccccccccc}
			\toprule
			\multirow{2}{*}{Methods}
			& \multicolumn{3}{c}{DIV2K-Real} & 
			& \multicolumn{3}{c}{2020Track2} &
			& \multicolumn{3}{c}{RealSRSet} \\ 
			\cmidrule{2-4} \cmidrule{6-8} \cmidrule{10-12} 
			& PSNR$\uparrow$     & SSIM$\uparrow$       & LPIPS$\downarrow$  &
			& NIQE$\downarrow$      & NRQM$\uparrow$& PI$\downarrow$     &
			& NIQE$\downarrow$      & NRQM$\uparrow$& PI$\downarrow$    \\
			\midrule
			BSRNet~\cite{bsrgan}
			&$25.12$&$0.6774$&$0.5113$&
			&$8.475$&$3.447$&$7.514$&  
			&$8.830$&$\mathbf{4.416}$&${6.807}$ \\
			Real-ESRNet~\cite{real-esrgan}
			&$24.75$&$0.6834$&$\mathbf{0.4917}$&
			&${8.083}$&${3.653}$&${7.215}$& 
			&${8.082}$&$4.300$&$6.892$\\
			DAN
			&$\mathbf{25.47}$&$\mathbf{0.6886}$&${0.5036}$&
			&$\mathbf{7.891}$&$\mathbf{3.673}$&$\mathbf{7.109}$& 
			&$\mathbf{7.367}$&${3.958}$&$\mathbf{6.704}$\\
			\midrule
			BSRGAN~\cite{bsrgan}
			&$24.28$&$0.6306$&$0.3861$&
			&$\mathbf{4.761}$&$6.164$&$4.300$&  
			&$5.468$&$6.254$&$4.607$ \\
			Real-ESRGAN~\cite{real-esrgan}
			&$23.45$&$0.6291$&$\mathbf{0.3527}$&
			&$5.114$&$5.600$&$4.758$& 
			&$5.615$&$5.988$&$4.813$\\
			Ji \etal (DPED)~\cite{ji2020real}
			&$20.96$&$0.4102$&$0.6374$&
			&$4.866$&$\mathbf{6.742}$&$\mathbf{4.062}$& 
			&$\mathbf{4.567}$&$\mathbf{6.473}$&$\mathbf{4.047}$\\
			DAN-GAN 
			&$\mathbf{24.91}$&$\mathbf{0.6591}$&$0.4069$&
			&$5.871$&$5.288$&$5.291$& 
			&$5.019$&$6.300$&$4.359$\\
			\bottomrule
	\end{tabular}}
\end{table*}
\begin{figure}[t]
	\centering
	\includegraphics[width=\linewidth]{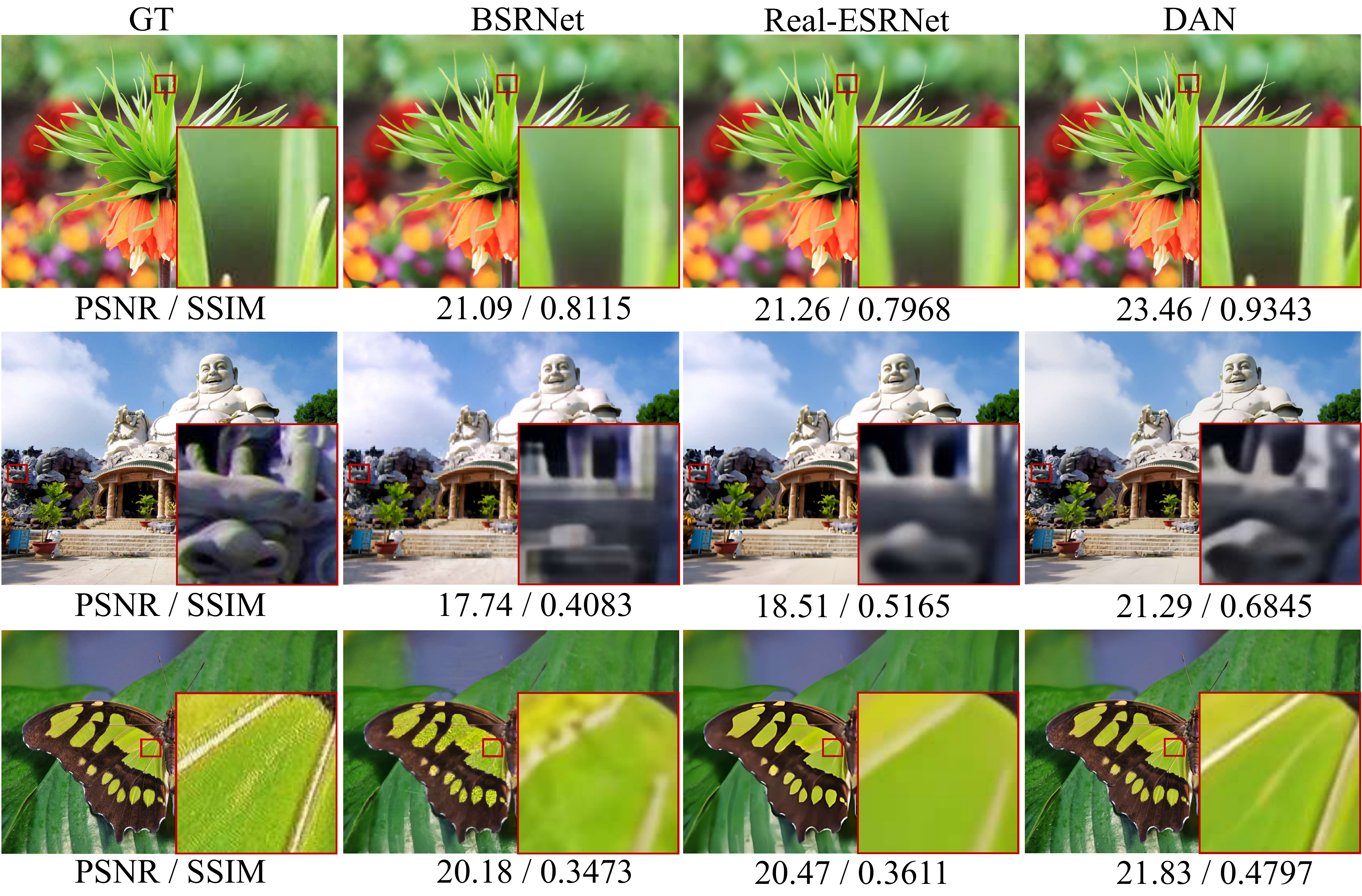}
	\caption{Visual comparisons between different methods on DIV2K-Real for scale factor $\times4$. Best viewed in color.}\label{fig:vis_setting2_syn}
\end{figure}

\begin{table*}[h]
	\centering
	\caption{Comparisons of complexity and speed of different methods. The best results are denoted in bold.} \label{tab:speed}
	\setlength{\tabcolsep}{24pt}
	\resizebox{\linewidth}{!}{
		\begin{tabular}{llll}
			\toprule
			Methods & \# Params (M) & Multi-Adds (G) & Speed (s/image) \\
			\midrule
			KernelGAN~\cite{kernel_gan}+ZSSR~\cite{zssr}
			&$0.30$& - &$120.27$ \\
			IKC~\cite{ikc}  
			&$5.29$&$2178.72$&$1.14$ \\
			DCLS~\cite{least_square}  
			&$13.63$&$368.15$&$0.18$ \\
			RealESRGAN~\cite{real-esrgan}  
			&$16.70$&$871.24$&$0.17$ \\
			DAN-Pre~\cite{dan}
			&$4.33$&$926.72$&$0.21$ \\
			DAN
			&$1.95$&$\mathbf{221.61}$&$\mathbf{0.05}$ \\
			\bottomrule
	\end{tabular}}
\end{table*}

\begin{figure*}[t]
	\centering
	\includegraphics[width=\linewidth]{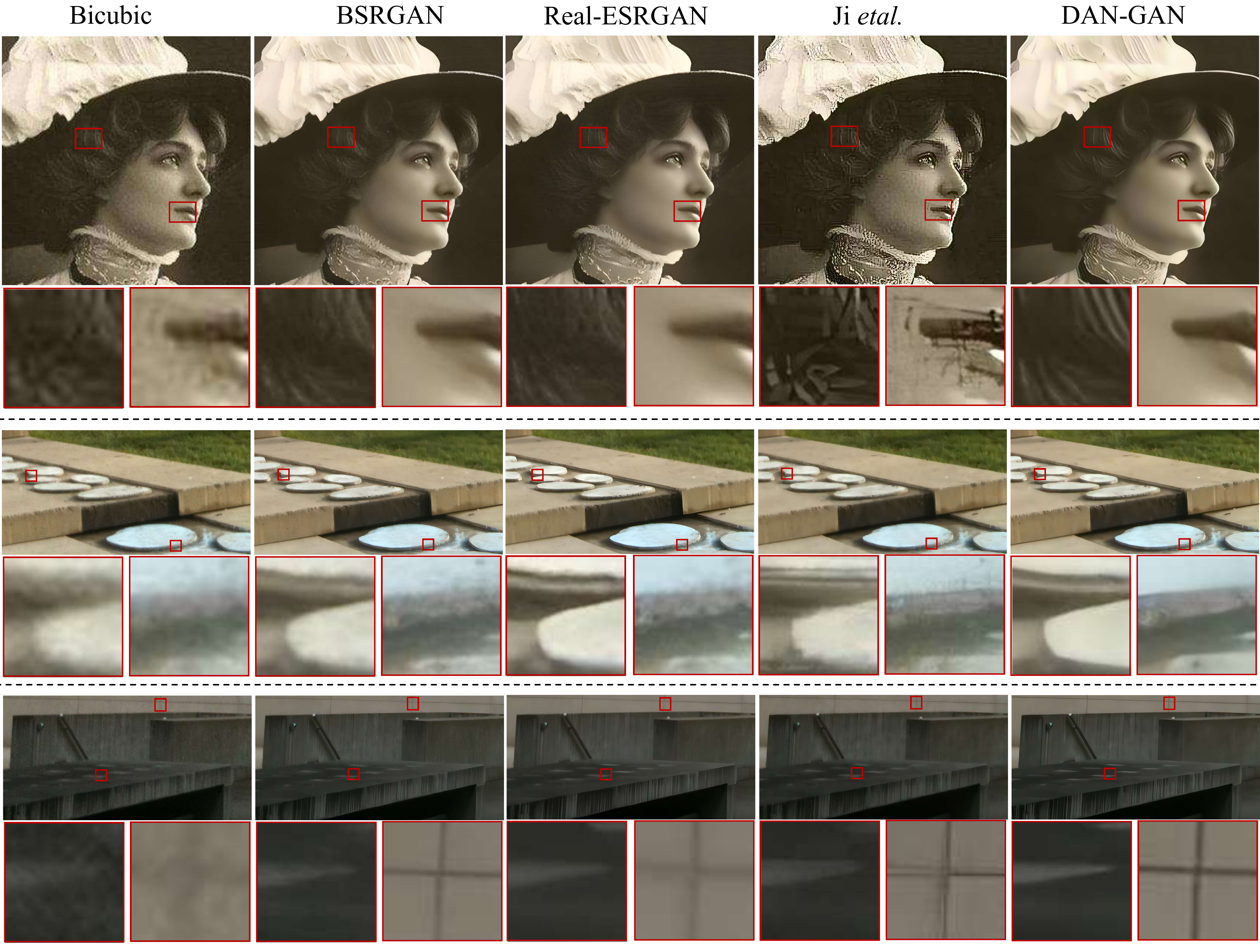}
	\caption{Visual comparisons between different methods on real images for scale factor $\times4$. Best  viewed in color.}\label{fig:vis_setting2_gan}
\end{figure*}

\subsection{Comparisons on Blurry LR images}\label{exp:compare_div2krk}
In this subsection, we explore the performance of DAN for blurry LR images. We mainly compare two types of methods: the two-step methods and end-to-end methods. A two-step method is usually the combination of a degradation estimation method and a non-blind SR method. Thus, we denote the two-step methods as 'A+B', where A is the degradation estimation method, and B is the non-blind SR method. In this section, we mainly compare with five kinds of combinations, including KernelGAN~\cite{kernel_gan}+ZSSR~\cite{zssr}, KernelGAN+SRMD~\cite{srmd}, KernelGAN+USRNet~\cite{usr}, Michaeli \etal~\cite{nonpara}+SRMD, and Michaeli \etal+ZSSR. For the end-to-end methods, we also mainly compare with five methods, \ie EDSR~\cite{edsr}, IKC~\cite{ikc}, AdaTarget~\cite{adatarget}, KOALAnet~\cite{koalanet}, and DCLS~\cite{least_square}. The result of bicubic interpolation is also added as a reference. We also include a comparison with the preliminary version of DAN, which is denoted as DAN-Pre. We need to note that the original EDSR is trained with bicubically synthesized samples. To make a comprehensive comparison, we retrain EDSR in the same way that we train DAN here. As we have discussed in Sec~\ref{sec:instantiation}, the \textit{Restorer} body in DAN can be chosen freely. To make the comparison with EDSR fair, we directly use the body of EDSR as the \textit{Restorer} body in DAN.

\vspace{0.02\linewidth}\noindent
\textbf{Quantitative comparisons.}
We evaluate the reference methods on DIV2KRK, and the results are shown in Table~\ref{tab:compare_div2krk}. USRNet~\cite{usr} and SRMD~\cite{srmd} are the leading non-blind SR methods. They perform excellently as long as the ground-truth blur kernels are provided. However, in real applications, the blur kernels are estimated, which may deviate from the ground-truth ones. Consequently, as shown in the table,  KernelGAN+USRNet and Michaeli \etal+SRMD can only achieve results even worse than simple bicubic interpolation ($20.06$ dB and $23.34$ dB V.S. $25.33$ dB for scale factor $\times4$). This may be because the two parts of the two-step method are separately optimized and thus they could not cooperate well with each other. As one can see from the table, most end-to-end methods perform better than all of the listed two-step methods, which strongly indicates the benefits of joint optimization.

Compared with KOALAnet and DCLS, which also adopt the end-to-end framework (proposed after DAN-Pre), DAN can still maintain its superiority (although DCLS achieves comparable results with DAN for scale factor $\times4$, it has a much larger model size than DAN, which will be discussed in Sec~\ref{sec:compare_speed}). It may be because the alternating optimization algorithm of DAN enables \textit{Restorer} and \textit{Estimator} to utilize the intermediate results of each other. This comparison also demonstrates the advantages of alternating optimization.

We also note that EDSR, which has a very simple network architecture (consists of only residual blocks), can perform better than most end-to-end methods. It indicates that the elaborated complex network architecture may not help improve the performance of blind SR methods. Inspired by the performance of EDSR, we also largely simplify the architecture of DAN-Pre. As shown in the table, the current version of DAN can perform much better than the preliminary version.  Additionally, the body of \textit{Restorer} in DAN has the same architecture as that of  EDSR. While DAN can largely outperform EDSR, which suggests that the \textit{Estimator} in DAN plays an important role in improving the performance of blind SR. EDSR super-resolves LR images regardless of their degradations. While \textit{Estimator} enables DAN to be degradation-aware and perform better over various degradations.

\vspace{0.02\linewidth}\noindent
\textbf{Qualitative comparisons.}
We visualize the $\times2$ SR results of \textit{img 823} and \textit{img 872} in Fig~\ref{fig:vis_setting1_x2}. As one can see, KerenelGAN+ZSSR and IKC fail to remove the blur and can only produce over-smoothed results. The SR images produced by DCLS are much sharper but contain some unpleasant artifacts, such as twisted lines. While the SR results of DAN are clearer, sharper, and contain fewer artifacts. The same comparisons are also shown in in Fig~\ref{fig:vis_setting1_x4}, which is the $\times4$ SR results of \textit{img 837} and \textit{img 845}. 

\subsection{Comparisons on Real-World LR Images}\label{sec:comapre_real}
In this section, we explore the performance of DAN for real-world LR images. As we have described above, we use the degradation model in Real-ESRGAN~\cite{real-esrgan} to synthesize training samples, which are then used to train DAN for real-world LR images. In this case, we mainly compare with three methods, \ie Ji \etal~\cite{ji2020real}, BSRGAN~\cite{bsrgan}, and Real-ESRGAN~\cite{real-esrgan}. These methods have two versions: the PSNR-oriented version and the GAN version. As the two versions have different behaviors, methods of different versions are compared independently. And we also provide the results of DAN and DAN-GAN.

\vspace{0.02\linewidth}\noindent
\textbf{Comparison results.}
We evaluate different methods on three datasets: the DIV2K-Real that we synthesized via the degradation model in~\cite{real-esrgan}, 2020Track2~\cite{ntire2020}, and the RealSRSet collected in~\cite{bsrgan}. Since there are no ground-truth for 2020Track2 and RealSRSet, we use the non-reference metrics for reference. As shown in Table~\ref{tab:compare_real}, DAN and DAN-GAN achieve the best PSNR and SSIM results on the synthetic dataset among the PSNR-oriented methods and perceptual-oriented methods respectively. Fig~\ref{fig:vis_setting2_syn} shows the visual comparisons on the synthetic dataset. As one can see, compared with BSRNet and Real-ESRNet, DAN produce can produce SR images with sharper and clearer textures. On 2020Track2 and RealSRSet,  DAN-GAN fail to achieve promising quantitative results. However, as discussed in~\cite{bsrgan,pdmsr}, these metrics may fail to measure the visual quality of SR images. As shown in Fig~\ref{fig:vis_setting2_gan},  although Ji \etal achieves the best quantitative results, its SR images contain serious artifacts, which are the worst in the reference methods. BSRGAN and Real-ESRGAN perform better, but the edges in their produced images are not clear enough. This may be because these methods are degradation-unaware, and are likely to produce over-smoothed images. As a comparison, DAN-GAN is degradation-specific, \ie performing SR according to the estimated degradation. As a result, DAN-GAN can produce sharper and clearer SR images.

\subsection{Comparisons on Complexity and Speed}\label{sec:compare_speed}

Compared with other blind SR methods, our end-to-end model also has superiority in model complexity inference speed. To make a quantitative comparison, we evaluate the average speed of different methods on the same platform with an RTX 3090 GPU. We choose the $\times4$ models of KernelGAN~\cite{kernel_gan} + ZSSR~\cite{zssr}, IKC~\cite{ikc}, and RealESRGAN~\cite{real-esrgan} as the comparison methods. The model complexity is measured by the number of parameters and multi-adds. And the speed is evaluated by the average time of processing $1000$ images. The multi-adds and speed are calculated when the size of their output images is $180\times270\times3$. The comparisons are shown in Table~\ref{tab:speed}. The number of multi-adds of KernelGAN+ZSSR is left out because it re-trains a different model for each test image. In that case, multi-adds can not indicate the model complexity. 

As one can see, the average speed of DAN-Pre is $0.21$ seconds per image,  nearly $554$ times faster than KernelGAN + ZSSR, and $5$ times faster than IKC, which demonstrates the speed superiority of our end-to-end framework. Moreover, in the current version, DAN is further simplified. Compared with DAN-Pre, DAN has $55\%$ fewer parameters and $77\%$ fewer multi-adds. The speed of DAN is also $4$ times faster than DAN-Pre. Compared with RealESRGAN, the current version of DAN also has $89\%$ fewer parameters, $74\%$ fewer multi-adds, and $3.4$ times faster speed. While as we have discussed in Sec~\ref{tab:compare_real}, DAN also performs better.

\subsection{Study of Estimated Degradations}\label{sec:deg_study}
\textbf{Accuracy.}
Previous degradation estimation methods mainly focus on estimating the blur kernel. Thus, to make better comparisons with other methods, we also evaluate the accuracy of estimated kernels in the case of SR for blurry LR images. We use two metrics to measure the accuracy: 1) we calculate the mean squared error (MSE) between the ground-truth kernel and the predicted kernel; 2) we degrade the original HR image with our predicted kernel and calculate the PSNR between the original LR image and our generated LR image. 

We choose four reference methods, \ie KernelGAN~\cite{kernel_gan}, CorrFilter~\cite{corrfilter}, DCLS~\cite{least_square}, and DAN-Pre~\cite{dan}.  The quantitative results are shown in Table~\ref{tab:kernel_acc}. It should be noted that the original version of DAN-Pre, as presented in our conference paper, can only estimate the PCA feature of the kernel instead of the whole kernel, whereas in this experiment, we have modified DAN-Pre to predict the entire kernel. As one can see, in terms of both Kernel-MSE and LP-PSNR, the estimation accuracy of DAN-Pre is much better than KernelGAN and CorrFilter.
This superiority could be attributed to two reasons: 
1) the \textit{Estimator} of DAN can utilize the information of intermediate SR results, which makes it easier for \textit{Estimator} to predict accurate degradation;
2) the \textit{Estimator} and \textit{Restorer} are optimized in an end-to-end network and are likely to get better compatible with each other.
DCLS, which is proposed after DAN-Pre and also adopts an end-to-end framework, achieves slightly better kernel-estimation performance than DAN-Pre. While the current version DAN superpasses it by a large margin. This may be because the \textit{Estimator} in DAN predicts the parameters ($\sigma_x$, $\sigma_y$, $\theta$, \etc) instead of the whole kernel, in which way the possible space the predicted kernel can be largely reduced, and the kernel estimation also becomes easier.
	
\begin{table*}[t]
	\centering
	\caption{Comparisons on the accuracy of estimated kernel. Results are calculated on the DIV2KRK (scale factor $\times4$) dataset. $\uparrow$ denotes the higher the better, and $\downarrow$ denotes the lower the better. The best result is denoted in bold.} \label{tab:kernel_acc}
	\setlength{\tabcolsep}{24pt}
	\resizebox{\linewidth}{!}{
		\begin{tabular}{lccccc}
			\toprule
			Methods &KernelGAN &CorrFilter& DCLS&DAN-Pre & DAN \\
			\midrule
			Kernel-MSE$\downarrow$
			&$0.1518$&$0.1392$&$0.0574$&$0.0817$&$\mathbf{8.104\times10^{-7}}$\\
			LR-PSNR$\uparrow$
			&$41.28$&$41.35$&$45.27$&$45.06$&$\mathbf{47.35}$\\
			\bottomrule
	\end{tabular}}
\end{table*}
\begin{figure}
	\centering
	\includegraphics[width=\linewidth]{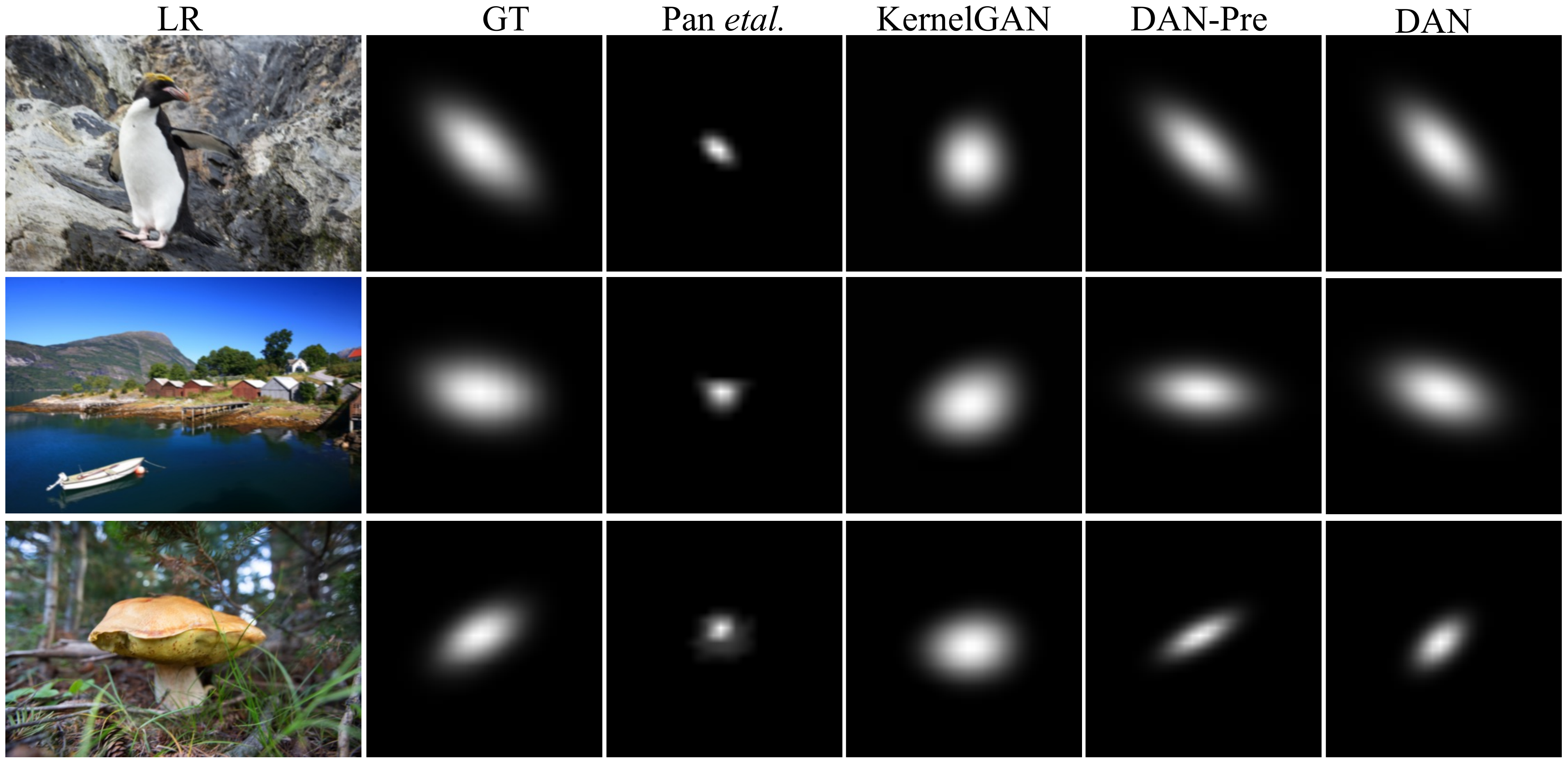}
	\caption{Visualization of  estimated blur kernels.}\label{vis_kernel}
\end{figure}

\vspace{0.02\linewidth}\noindent
\textbf{Visualization.}
we also visualize some estimated kernels on DIV2KRK (scale factor $\times4$) to qualitatively measure the performance of \textit{Estimator}.
We use the results of  KernelGAN~\cite{kernel_gan}, Pan \etal~\cite{pan2014}, DAN-Pre~\cite{dan} as comparisons. As shown in Fig~\ref{vis_kernel},  the kernels estimated by KernelGAN are likely to be isotropic and look very different from the ground-truth kernels. Instead, both DAN-Pre and DAN can estimate the kernel much more accurately, even if the ground-truth kernels are highly anisotropic. 

\subsection{Non-blind Setting}
To explore the influence of the degradation-estimation deviation, we replace the estimated degradations with the ground-truth ones and observe the performance of the \textit{Restorer}. The comparisons are shown in Table~\ref{tab:non_blind_setting}. As one can see, when the ground-truth degradations are provided, the results of DAN almost keep unchanged. It indicates that \textit{Restorer} is not sensitive to estimation deviation of the degradation. This is because the \textit{Estimator} and the \textit{Restorer} are jointly optimized and they can be more tolerant of the estimation deviations of each other. The superiority of DAN also partially comes from the good cooperation between its \textit{Estimator} and \textit{Restorer}.

\begin{table}[t]
	\centering
	\caption{Average results of DAN on the DIV2K-Real dataset. $\uparrow$ denotes the higher the better, and $\downarrow$ denotes the lower the better.} \label{tab:non_blind_setting}
	\setlength{\tabcolsep}{12pt}
	\resizebox{\linewidth}{!}{
		\begin{tabular}{lccc}
			\toprule
			& PSNR$\uparrow$ & SSIM$\uparrow$ & LPIPS$\downarrow$ \\
			\midrule
			w/ GT  
			&$25.47$&$0.6885$&$0.5008$ \\
			w/o GT  
			&$25.47$&$0.6886$&$0.5036$\\
			\bottomrule
	\end{tabular}}
\end{table}

\subsection{Study of Iterations}\label{sec:iter_exp}

In this section, we experimentally explore the influence of the iterations of the alternating optimization. We gradually change the iterations from $1$ to $4$ and train DAN with different iterations. The quantitative results on DIV2K-Real are shown in Table~\ref{tab:study_iter}. As one can see, as the number of iterations increases, the performance of DAN also grows monotonically. However, more iterations also require a longer inference time. We finally set the number of alternating iterations in DAN as $3$ for the balance of cost and effectiveness. 

\begin{figure}[t]
	\centering
	\includegraphics[width=\linewidth]{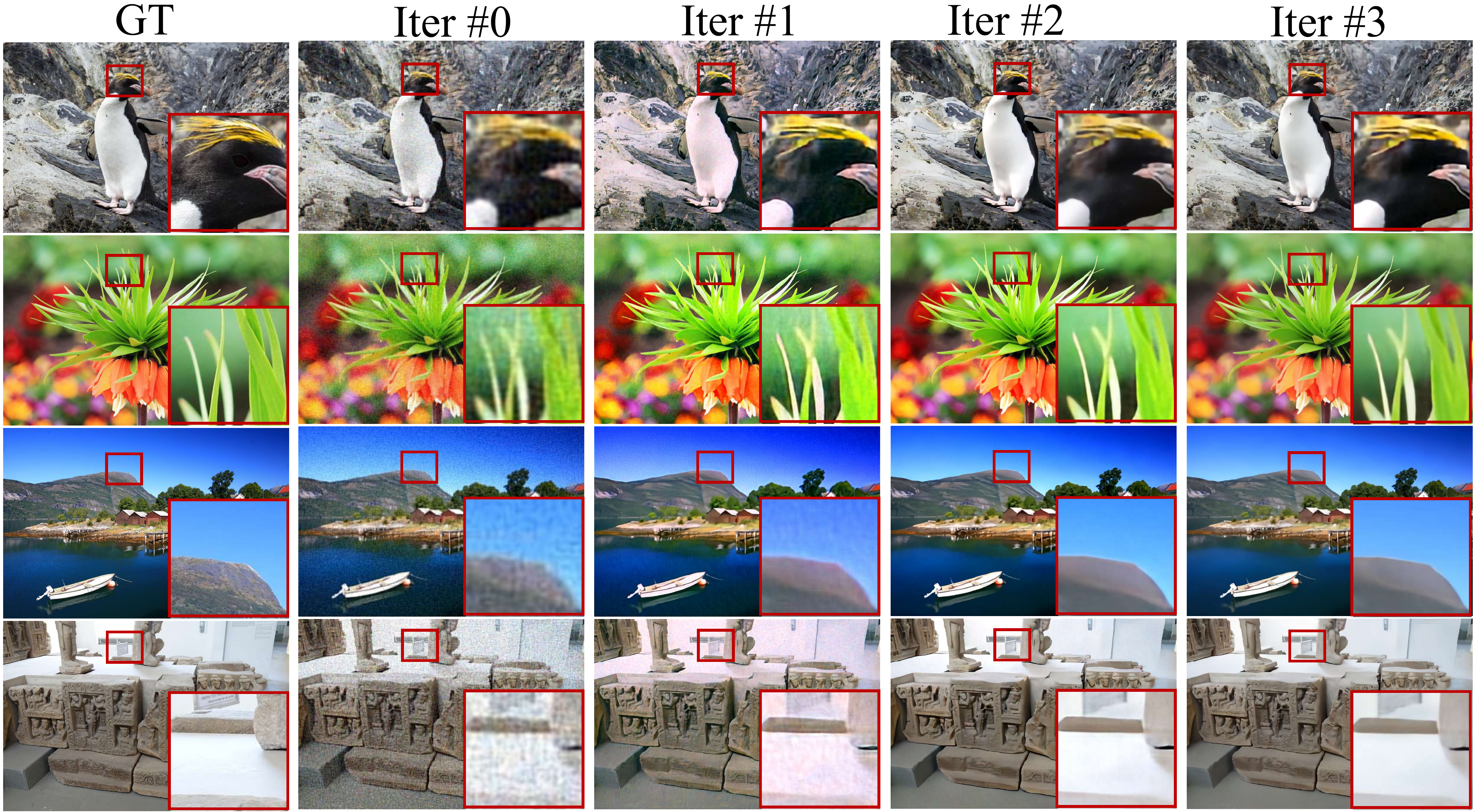}
	\caption{Visualization of the intermediate SR images during the alternating process.}\label{fig:iter_process}
\end{figure}

We also change the iterations of a trained DAN to explore its iterating process. It should be noted that the tail modules in DAN are only trained to process the features at the last iterations. If we directly change the iterations in a trained DAN, the tail modules need to process features at intermediate iterations, which may be a different domain of features at the last one. Thus, to comprehensively explore the performance of intermediate iterations, we train tail modules for each iteration respectively, while keeping the weights of DAN fixed, in which case, the intermediate performance can be better evaluated. As shown in Fig~\ref{fig:iter_process}, the SR results exhibit improved visual quality with cleaner images and richer details as the iterations progress. We also show the quantitative results in Table~\ref{tab:trained_iter}. 

\begin{table}[t]
	\centering
	\caption{The quantitative results of DAN trained with different number of  alternating iterations. Average results are reported on the DIV2K-Real dataset. $\uparrow$ denotes the higher the better, and $\downarrow$ denotes the lower the better.} \label{tab:study_iter}
	\setlength{\tabcolsep}{12pt}
	\resizebox{\linewidth}{!}{
		\begin{tabular}{cccc}
			\toprule
			\# Iters & PSNR$\uparrow$ & SSIM$\uparrow$ & LPIPS$\downarrow$ \\
			\midrule
			1 
			&$25.20$&$0.6822$&$0.5104$ \\
			2  
			&$25.33$&$0.6840$&$0.5088$ \\
			3
			&$25.47$&$0.6886$&$0.5036$  \\
			4
			&$25.53$&$0.6904$&$0.5021$ \\
			\bottomrule
	\end{tabular}}
\end{table}

\begin{table}[t]
	\centering
	\caption{The quantitative results of different alternating iterations in a trained DAN. $\uparrow$ indicates the higher the better, and $\downarrow$ indicates the lower the better.}\label{tab:trained_iter}
	\setlength{\tabcolsep}{10pt}
	\resizebox{\linewidth}{!}{
		\begin{tabular}{ccccc}
			\toprule
			\# Iters & 0 & 1 & 2 & 3 \\
			\midrule
			PSNR$\uparrow$ & $21.60$ & $24.14$ & $25.10$&$25.47$\\
			SSIM$\uparrow$ & $0.6115$ & $0.6347$ &$0.6737$&$0.6886$\\
			LPIPS$\downarrow$  & $0.7138$ & $0.7085$ &$0.5391$& $0.5036$\\
			\bottomrule
		\end{tabular}
	}
\end{table}

\subsection{Ablation Studies about the Network Architecture}
In this section, we perform ablation studies to validate the the advantages of current version against the preliminary version~\cite{dan}, \ie DAN-Pre. There are three main modifications, including A) concatenating the two inputs of  \textit{Estimator} and \textit{Restorer} at the beginning; B) iterating in the feature space; and C) setting the initial degradation learnable. We use the preliminary version as the baseline method. Then we apply the three modifications in turn to the baseline method and compare the results. As shown in Table~\ref{tab:aba_arch}, if the modification A is applied,cite the number of parameters can be largely reduced, while the performance becomes better instead. It suggests that the conditional residual block in DAN-Pre may have the only limited representing capacity and consume too many unnecessary parameters. If modification B is further applied, the performance of can be further improved (the PSNR result increases from $24.98$ dB to $25.24$ dB). It may be attributed to that iterating in the feature space can get the information better transferred among different iterations. And the modification C, \ie setting the initial degradation learnable, improves the PSNR result from $25.24$ dB to $25.41$ dB. It suggests that a learnable initial state may make it easier for DAN to converge to a better point.

\begin{table}[h]
	\centering
	\caption{Ablation studies on the network architectures. Average results are reported on the DIV2K-Real dataset. $\uparrow$ denotes the higher the better, and $\downarrow$ denotes the lower the better. The best two results are indicated in bold and underlined respectively.} \label{tab:aba_arch}
	\setlength{\tabcolsep}{8pt}
	\resizebox{\linewidth}{!}{
		\begin{tabular}{ccccccc}
			\toprule
			A & B & C & \# Params (M)&PSNR$\uparrow$ & SSIM$\uparrow$ & LPIPS$\downarrow$ \\
			\midrule
			&& 
			&$4.33$&$24.33$&$0.6712$&$0.5251$ \\
			\checkmark & & 
			&$1.94$&$24.98$&$0.6821$&$0.5121$\\
			\checkmark & \checkmark &
			&$1.94$&$25.24$&$0.6830$&$0.5090$ \\
			\checkmark & \checkmark &\checkmark
			&$1.95$&$25.47$&$0.6886$&$0.5036$\\
			\bottomrule
	\end{tabular}}
\end{table}

\subsection{Study of the Loss Function for \textit{Estimator}}
In this paper, the estimation of degradation parameters is formulated as a regression task and is supervised using L2 loss.  While some degradation parameters ( \eg $k_g$, $k_c$, $n_t$ \etc) are binary-valued. The estimation of these parameters actually can be formulated as a classification task and supervised using cross-entropy (CE) loss. In this section, we partitions the degradation parameters into two categories: discrete-values (\eg, kernel types, noise types, scaling modes, \etc) and continuous-values (\eg, kernel radius, noise level, quality factor, \etc). The discrete-value part is supervised using CE loss, while the continuous-value part is still supervised using L2 loss. As shown in Table~\ref{tab:loss}, the performance of DAN is comparable with both types of losses. However, when a mixture of CE loss and L2 loss is used, slightly better SSIM and LPIPS scores are obtained. Nevertheless, implementing L2 loss alone is significantly more straightforward compared to a mixed loss. As a result, we have decided to use only L2 loss.

\begin{table}[t]
	\centering
	\caption{The comparison between different loss types for \textit{Estimator}.$\uparrow$ indicates the higher the better, and $\downarrow$ indicates the lower the better.}\label{tab:loss}
	\setlength{\tabcolsep}{8pt}
	\resizebox{\linewidth}{!}{
		\begin{tabular}{lccc}
			\toprule
			Loss Types & PSNR$\uparrow$&SSIM$\uparrow$&LPIPS$\downarrow$ \\
			\midrule
			Mixture of CE and L2 loss &$25.47$&$0.6887$&$0.5021$\\
			Only L2 loss&$25.47$&$0.6886$&$0.5036$\\
			\bottomrule
		\end{tabular}
	}
\end{table}

\section{Conclusion}

In this paper, we have proposed an end-to-end algorithm that can simultaneously estimate the complex degradations and restore the SR images in blind SR. This algorithm is based on alternating optimization and consists of two parts, namely \textit{Restorer} and \textit{Estimator}. We implement the two parts by convolutional modules and unfold the alternating process to form an end-to-end trainable network. In this way, \textit{Estimator} can utilize information from intermediate SR images, which makes it easier to estimate the degradation. More importantly, \textit{Restorer} is trained with the degradations estimated by \textit{Estimator}, instead of the ground-truth ones. Thus \textit{Restorer} could be more tolerant to the estimation error of \textit{Estimator}. Experiments show that the well-compatibility of the two modules can largely improve the accuracy of blind SR, which demonstrates the importance of an end-to-end pipeline. Moreover, compared with those degradation-unaware methods, the proposed method performs SR according to the degradations of LR images and is likely to produce sharper and clearer SR images.


{
	\bibliography{egbib}


\begin{thebibliography}{78}
\ifx \bisbn   \undefined \def \bisbn  #1{ISBN #1}\fi
\ifx \binits  \undefined \def \binits#1{#1}\fi
\ifx \bauthor  \undefined \def \bauthor#1{#1}\fi
\ifx \batitle  \undefined \def \batitle#1{#1}\fi
\ifx \bjtitle  \undefined \def \bjtitle#1{#1}\fi
\ifx \bvolume  \undefined \def \bvolume#1{\textbf{#1}}\fi
\ifx \byear  \undefined \def \byear#1{#1}\fi
\ifx \bissue  \undefined \def \bissue#1{#1}\fi
\ifx \bfpage  \undefined \def \bfpage#1{#1}\fi
\ifx \blpage  \undefined \def \blpage #1{#1}\fi
\ifx \burl  \undefined \def \burl#1{\textsf{#1}}\fi
\ifx \doiurl  \undefined \def \doiurl#1{\url{https://doi.org/#1}}\fi
\ifx \betal  \undefined \def \betal{\textit{et al.}}\fi
\ifx \binstitute  \undefined \def \binstitute#1{#1}\fi
\ifx \binstitutionaled  \undefined \def \binstitutionaled#1{#1}\fi
\ifx \bctitle  \undefined \def \bctitle#1{#1}\fi
\ifx \beditor  \undefined \def \beditor#1{#1}\fi
\ifx \bpublisher  \undefined \def \bpublisher#1{#1}\fi
\ifx \bbtitle  \undefined \def \bbtitle#1{#1}\fi
\ifx \bedition  \undefined \def \bedition#1{#1}\fi
\ifx \bseriesno  \undefined \def \bseriesno#1{#1}\fi
\ifx \blocation  \undefined \def \blocation#1{#1}\fi
\ifx \bsertitle  \undefined \def \bsertitle#1{#1}\fi
\ifx \bsnm \undefined \def \bsnm#1{#1}\fi
\ifx \bsuffix \undefined \def \bsuffix#1{#1}\fi
\ifx \bparticle \undefined \def \bparticle#1{#1}\fi
\ifx \barticle \undefined \def \barticle#1{#1}\fi
\bibcommenthead
\ifx \bconfdate \undefined \def \bconfdate #1{#1}\fi
\ifx \botherref \undefined \def \botherref #1{#1}\fi
\ifx \url \undefined \def \url#1{\textsf{#1}}\fi
\ifx \bchapter \undefined \def \bchapter#1{#1}\fi
\ifx \bbook \undefined \def \bbook#1{#1}\fi
\ifx \bcomment \undefined \def \bcomment#1{#1}\fi
\ifx \oauthor \undefined \def \oauthor#1{#1}\fi
\ifx \citeauthoryear \undefined \def \citeauthoryear#1{#1}\fi
\ifx \endbibitem  \undefined \def \endbibitem {}\fi
\ifx \bconflocation  \undefined \def \bconflocation#1{#1}\fi
\ifx \arxivurl  \undefined \def \arxivurl#1{\textsf{#1}}\fi
\csname PreBibitemsHook\endcsname

\bibitem{usr}
\begin{bchapter}
\bauthor{\bsnm{Zhang}, \binits{K.}},
\bauthor{\bsnm{Gool}, \binits{L.V.}},
\bauthor{\bsnm{Timofte}, \binits{R.}}:
\bctitle{Deep unfolding network for image super-resolution}.
In: \bbtitle{Proceedings of the IEEE/CVF Conference on Computer Vision and
  Pattern Recognition},
pp. \bfpage{3217}--\blpage{3226}
(\byear{2020})
\end{bchapter}
\endbibitem

\bibitem{baker2002limits}
\begin{barticle}
\bauthor{\bsnm{Baker}, \binits{S.}},
\bauthor{\bsnm{Kanade}, \binits{T.}}:
\batitle{Limits on super-resolution and how to break them}.
\bjtitle{IEEE Transactions on Pattern Analysis and Machine Intelligence}
\bvolume{24}(\bissue{9}),
\bfpage{1167}--\blpage{1183}
(\byear{2002})
\end{barticle}
\endbibitem

\bibitem{nonpara}
\begin{botherref}
\oauthor{\bsnm{Michaeli}, \binits{T.}},
\oauthor{\bsnm{Irani}, \binits{M.}}:
Nonparametric blind super-resolution.
IEEE International Conference on Computer Vision,
945--952
(2013)
\end{botherref}
\endbibitem

\bibitem{kernel_gan}
\begin{bchapter}
\bauthor{\bsnm{Bell-Kligler}, \binits{S.}},
\bauthor{\bsnm{Shocher}, \binits{A.}},
\bauthor{\bsnm{Irani}, \binits{M.}}:
\bctitle{Blind super-resolution kernel estimation using an internal-gan}.
In: \bbtitle{Advances in Neural Information Processing Systems}
(\byear{2019})
\end{bchapter}
\endbibitem

\bibitem{dpsr}
\begin{bchapter}
\bauthor{\bsnm{{Kai Zhang and Wangmeng Zuo and Lei Zhang}}}:
\bctitle{Deep plug-and-play super-resolution for arbitrary blur kernels}.
In: \bbtitle{Proceedings of the IEEE/CVF Conference on Computer Vision and
  Pattern Recognition},
pp. \bfpage{1671}--\blpage{1681}
(\byear{2019})
\end{bchapter}
\endbibitem

\bibitem{least_square}
\begin{bchapter}
\bauthor{\bsnm{Luo}, \binits{Z.}},
\bauthor{\bsnm{Huang}, \binits{H.}},
\bauthor{\bsnm{Yu}, \binits{L.}},
\bauthor{\bsnm{Li}, \binits{Y.}},
\bauthor{\bsnm{Fan}, \binits{H.}},
\bauthor{\bsnm{Liu}, \binits{S.}}:
\bctitle{Deep constrained least squares for blind image super-resolution}.
In: \bbtitle{Proceedings of the IEEE/CVF Conference on Computer Vision and
  Pattern Recognition}
(\byear{2022})
\end{bchapter}
\endbibitem

\bibitem{dan}
\begin{botherref}
\oauthor{\bsnm{Luo}, \binits{Z.}},
\oauthor{\bsnm{Huang}, \binits{Y.}},
\oauthor{\bsnm{Li}, \binits{S.}},
\oauthor{\bsnm{Wang}, \binits{L.}},
\oauthor{\bsnm{Tan}, \binits{T.}}:
Unfolding the alternating optimization for blind super resolution.
Advances in Neural Information Processing Systems
\textbf{33}
(2020)
\end{botherref}
\endbibitem

\bibitem{rcan}
\begin{bchapter}
\bauthor{\bsnm{Zhang}, \binits{Y.}},
\bauthor{\bsnm{Li}, \binits{K.}},
\bauthor{\bsnm{Li}, \binits{K.}},
\bauthor{\bsnm{Wang}, \binits{L.}},
\bauthor{\bsnm{Zhong}, \binits{B.}},
\bauthor{\bsnm{Fu}, \binits{Y.}}:
\bctitle{Image super-resolution using very deep residual channel attention
  networks}.
In: \bbtitle{Proceedings of the European Conference on Computer Vision},
pp. \bfpage{286}--\blpage{301}
(\byear{2018})
\end{bchapter}
\endbibitem

\bibitem{esrgan}
\begin{bchapter}
\bauthor{\bsnm{Wang}, \binits{X.}},
\bauthor{\bsnm{Yu}, \binits{K.}},
\bauthor{\bsnm{Wu}, \binits{S.}},
\bauthor{\bsnm{Gu}, \binits{J.}},
\bauthor{\bsnm{Liu}, \binits{Y.}},
\bauthor{\bsnm{Dong}, \binits{C.}},
\bauthor{\bsnm{Qiao}, \binits{Y.}},
\bauthor{\bsnm{Change~Loy}, \binits{C.}}:
\bctitle{Esrgan: Enhanced super-resolution generative adversarial networks}.
In: \bbtitle{Proceedings of the European Conference on Computer Vision
  Workshops},
pp. \bfpage{0}--\blpage{0}
(\byear{2018})
\end{bchapter}
\endbibitem

\bibitem{swinir}
\begin{bchapter}
\bauthor{\bsnm{Liang}, \binits{J.}},
\bauthor{\bsnm{Cao}, \binits{J.}},
\bauthor{\bsnm{Sun}, \binits{G.}},
\bauthor{\bsnm{Zhang}, \binits{K.}},
\bauthor{\bsnm{Van~Gool}, \binits{L.}},
\bauthor{\bsnm{Timofte}, \binits{R.}}:
\bctitle{Swinir: Image restoration using swin transformer}.
In: \bbtitle{Proceedings of the IEEE/CVF International Conference on Computer
  Vision},
pp. \bfpage{1833}--\blpage{1844}
(\byear{2021})
\end{bchapter}
\endbibitem

\bibitem{survey}
\begin{botherref}
\oauthor{\bsnm{Wang}, \binits{Z.}},
\oauthor{\bsnm{Chen}, \binits{J.}},
\oauthor{\bsnm{Hoi}, \binits{S.C.}}:
Deep learning for image super-resolution: A survey.
IEEE transactions on pattern analysis and machine intelligence
(2020)
\end{botherref}
\endbibitem

\bibitem{zhang2022memory}
\begin{barticle}
\bauthor{\bsnm{Zhang}, \binits{H.}},
\bauthor{\bsnm{Li}, \binits{Y.}},
\bauthor{\bsnm{Chen}, \binits{H.}},
\bauthor{\bsnm{Gong}, \binits{C.}},
\bauthor{\bsnm{Bai}, \binits{Z.}},
\bauthor{\bsnm{Shen}, \binits{C.}}:
\batitle{Memory-efficient hierarchical neural architecture search for image
  restoration}.
\bjtitle{International Journal of Computer Vision}
\bvolume{130}(\bissue{1}),
\bfpage{157}--\blpage{178}
(\byear{2022})
\end{barticle}
\endbibitem

\bibitem{zhou2022memory}
\begin{botherref}
\oauthor{\bsnm{Zhou}, \binits{M.}},
\oauthor{\bsnm{Yan}, \binits{K.}},
\oauthor{\bsnm{Pan}, \binits{J.}},
\oauthor{\bsnm{Ren}, \binits{W.}},
\oauthor{\bsnm{Xie}, \binits{Q.}},
\oauthor{\bsnm{Cao}, \binits{X.}}:
Memory-augmented deep unfolding network for guided image super-resolution.
International Journal of Computer Vision
(2022)
\end{botherref}
\endbibitem

\bibitem{sc}
\begin{barticle}
\bauthor{\bsnm{Yang}, \binits{J.}},
\bauthor{\bsnm{Wright}, \binits{J.}},
\bauthor{\bsnm{Huang}, \binits{T.S.}},
\bauthor{\bsnm{Ma}, \binits{Y.}}:
\batitle{Image super-resolution via sparse representation}.
\bjtitle{IEEE transactions on image processing}
\bvolume{19}(\bissue{11}),
\bfpage{2861}--\blpage{2873}
(\byear{2010})
\end{barticle}
\endbibitem

\bibitem{kk}
\begin{barticle}
\bauthor{\bsnm{Kim}, \binits{K.I.}},
\bauthor{\bsnm{Kwon}, \binits{Y.}}:
\batitle{Single-image super-resolution using sparse regression and natural
  image prior}.
\bjtitle{IEEE transactions on pattern analysis and machine intelligence}
\bvolume{32}(\bissue{6}),
\bfpage{1127}--\blpage{1133}
(\byear{2010})
\end{barticle}
\endbibitem

\bibitem{a+}
\begin{bchapter}
\bauthor{\bsnm{Timofte}, \binits{R.}},
\bauthor{\bsnm{De~Smet}, \binits{V.}},
\bauthor{\bsnm{Van~Gool}, \binits{L.}}:
\bctitle{Anchored neighborhood regression for fast example-based
  super-resolution}.
In: \bbtitle{Proceedings of the IEEE International Conference on Computer
  Vision},
pp. \bfpage{1920}--\blpage{1927}
(\byear{2013})
\end{bchapter}
\endbibitem

\bibitem{srcnn}
\begin{barticle}
\bauthor{\bsnm{Dong}, \binits{C.}},
\bauthor{\bsnm{Loy}, \binits{C.C.}},
\bauthor{\bsnm{He}, \binits{K.}},
\bauthor{\bsnm{Tang}, \binits{X.}}:
\batitle{Image super-resolution using deep convolutional networks}.
\bjtitle{IEEE Transactions on Pattern Analysis and Machine Intelligence}
\bvolume{38}(\bissue{2}),
\bfpage{295}--\blpage{307}
(\byear{2015})
\end{barticle}
\endbibitem

\bibitem{dbpn}
\begin{bchapter}
\bauthor{\bsnm{Haris}, \binits{M.}},
\bauthor{\bsnm{Shakhnarovich}, \binits{G.}},
\bauthor{\bsnm{Ukita}, \binits{N.}}:
\bctitle{Deep back-projection networks for super-resolution}.
In: \bbtitle{Proceedings of the IEEE/CVF Conference on Computer Vision and
  Pattern Recognition},
pp. \bfpage{1664}--\blpage{1673}
(\byear{2018})
\end{bchapter}
\endbibitem

\bibitem{meta_sr}
\begin{bchapter}
\bauthor{\bsnm{Hu}, \binits{X.}},
\bauthor{\bsnm{Mu}, \binits{H.}},
\bauthor{\bsnm{Zhang}, \binits{X.}},
\bauthor{\bsnm{Wang}, \binits{Z.}},
\bauthor{\bsnm{Tan}, \binits{T.}},
\bauthor{\bsnm{Sun}, \binits{J.}}:
\bctitle{Meta-sr: A magnification-arbitrary network for super-resolution}.
In: \bbtitle{Proceedings of the IEEE/CVF Conference on Computer Vision and
  Pattern Recognition},
pp. \bfpage{1575}--\blpage{1584}
(\byear{2019})
\end{bchapter}
\endbibitem

\bibitem{imdn}
\begin{bchapter}
\bauthor{\bsnm{Hui}, \binits{Z.}},
\bauthor{\bsnm{Gao}, \binits{X.}},
\bauthor{\bsnm{Yang}, \binits{Y.}},
\bauthor{\bsnm{Wang}, \binits{X.}}:
\bctitle{Lightweight image super-resolution with information multi-distillation
  network}.
In: \bbtitle{Proceedings of the 27th ACM International Conference on
  Multimedia},
pp. \bfpage{2024}--\blpage{2032}
(\byear{2019})
\end{bchapter}
\endbibitem

\bibitem{carn}
\begin{bchapter}
\bauthor{\bsnm{Ahn}, \binits{N.}},
\bauthor{\bsnm{Kang}, \binits{B.}},
\bauthor{\bsnm{Sohn}, \binits{K.-A.}}:
\bctitle{Fast, accurate, and lightweight super-resolution with cascading
  residual network}.
In: \bbtitle{Proceedings of the European Conference on Computer Vision},
pp. \bfpage{252}--\blpage{268}
(\byear{2018})
\end{bchapter}
\endbibitem

\bibitem{idn}
\begin{bchapter}
\bauthor{\bsnm{Hui}, \binits{Z.}},
\bauthor{\bsnm{Wang}, \binits{X.}},
\bauthor{\bsnm{Gao}, \binits{X.}}:
\bctitle{Fast and accurate single image super-resolution via information
  distillation network}.
In: \bbtitle{Proceedings of the IEEE/CVF Conference on Computer Vision and
  Pattern Recognition},
pp. \bfpage{723}--\blpage{731}
(\byear{2018})
\end{bchapter}
\endbibitem

\bibitem{fsrcnn}
\begin{bchapter}
\bauthor{\bsnm{Dong}, \binits{C.}},
\bauthor{\bsnm{Loy}, \binits{C.C.}},
\bauthor{\bsnm{Tang}, \binits{X.}}:
\bctitle{Accelerating the super-resolution convolutional neural network}.
In: \bbtitle{Proceedings of the European Conference on Computer Vision},
pp. \bfpage{391}--\blpage{407}
(\byear{2016}).
\bcomment{Springer}
\end{bchapter}
\endbibitem

\bibitem{vdsr}
\begin{bchapter}
\bauthor{\bsnm{Kim}, \binits{J.}},
\bauthor{\bsnm{Kwon~Lee}, \binits{J.}},
\bauthor{\bsnm{Mu~Lee}, \binits{K.}}:
\bctitle{Accurate image super-resolution using very deep convolutional
  networks}.
In: \bbtitle{Proceedings of the IEEE/CVF Conference on Computer Vision and
  Pattern Recognition},
pp. \bfpage{1646}--\blpage{1654}
(\byear{2016})
\end{bchapter}
\endbibitem

\bibitem{pixel_shuffle}
\begin{bchapter}
\bauthor{\bsnm{Shi}, \binits{W.}},
\bauthor{\bsnm{Caballero}, \binits{J.}},
\bauthor{\bsnm{Husz{\'a}r}, \binits{F.}},
\bauthor{\bsnm{Totz}, \binits{J.}},
\bauthor{\bsnm{Aitken}, \binits{A.P.}},
\bauthor{\bsnm{Bishop}, \binits{R.}},
\bauthor{\bsnm{Rueckert}, \binits{D.}},
\bauthor{\bsnm{Wang}, \binits{Z.}}:
\bctitle{Real-time single image and video super-resolution using an efficient
  sub-pixel convolutional neural network}.
In: \bbtitle{Proceedings of the IEEE/CVF Conference on Computer Vision and
  Pattern Recognition},
pp. \bfpage{1874}--\blpage{1883}
(\byear{2016})
\end{bchapter}
\endbibitem

\bibitem{san}
\begin{bchapter}
\bauthor{\bsnm{Dai}, \binits{T.}},
\bauthor{\bsnm{Cai}, \binits{J.}},
\bauthor{\bsnm{Zhang}, \binits{Y.}},
\bauthor{\bsnm{Xia}, \binits{S.-T.}},
\bauthor{\bsnm{Zhang}, \binits{L.}}:
\bctitle{Second-order attention network for single image super-resolution}.
In: \bbtitle{Proceedings of the IEEE/CVF Conference on Computer Vision and
  Pattern Recognition},
pp. \bfpage{11065}--\blpage{11074}
(\byear{2019})
\end{bchapter}
\endbibitem

\bibitem{transformer}
\begin{botherref}
\oauthor{\bsnm{Vaswani}, \binits{A.}},
\oauthor{\bsnm{Shazeer}, \binits{N.}},
\oauthor{\bsnm{Parmar}, \binits{N.}},
\oauthor{\bsnm{Uszkoreit}, \binits{J.}},
\oauthor{\bsnm{Jones}, \binits{L.}},
\oauthor{\bsnm{Gomez}, \binits{A.N.}},
\oauthor{\bsnm{Kaiser}, \binits{{\L}.}},
\oauthor{\bsnm{Polosukhin}, \binits{I.}}:
Attention is all you need.
Advances in neural information processing systems
\textbf{30}
(2017)
\end{botherref}
\endbibitem

\bibitem{ipt}
\begin{bchapter}
\bauthor{\bsnm{Chen}, \binits{H.}},
\bauthor{\bsnm{Wang}, \binits{Y.}},
\bauthor{\bsnm{Guo}, \binits{T.}},
\bauthor{\bsnm{Xu}, \binits{C.}},
\bauthor{\bsnm{Deng}, \binits{Y.}},
\bauthor{\bsnm{Liu}, \binits{Z.}},
\bauthor{\bsnm{Ma}, \binits{S.}},
\bauthor{\bsnm{Xu}, \binits{C.}},
\bauthor{\bsnm{Xu}, \binits{C.}},
\bauthor{\bsnm{Gao}, \binits{W.}}:
\bctitle{Pre-trained image processing transformer}.
In: \bbtitle{Proceedings of the IEEE/CVF Conference on Computer Vision and
  Pattern Recognition},
pp. \bfpage{12299}--\blpage{12310}
(\byear{2021})
\end{bchapter}
\endbibitem

\bibitem{bridge}
\begin{barticle}
\bauthor{\bsnm{K{\"o}hler}, \binits{T.}},
\bauthor{\bsnm{B{\"a}tz}, \binits{M.}},
\bauthor{\bsnm{Naderi}, \binits{F.}},
\bauthor{\bsnm{Kaup}, \binits{A.}},
\bauthor{\bsnm{Maier}, \binits{A.}},
\bauthor{\bsnm{Riess}, \binits{C.}}:
\batitle{Toward bridging the simulated-to-real gap: Benchmarking
  super-resolution on real data}.
\bjtitle{IEEE transactions on pattern analysis and machine intelligence}
\bvolume{42}(\bissue{11}),
\bfpage{2944}--\blpage{2959}
(\byear{2019})
\end{barticle}
\endbibitem

\bibitem{cycleSR}
\begin{bchapter}
\bauthor{\bsnm{Chen}, \binits{S.}},
\bauthor{\bsnm{Han}, \binits{Z.}},
\bauthor{\bsnm{Dai}, \binits{E.}},
\bauthor{\bsnm{Jia}, \binits{X.}},
\bauthor{\bsnm{Liu}, \binits{Z.}},
\bauthor{\bsnm{Xing}, \binits{L.}},
\bauthor{\bsnm{Zou}, \binits{X.}},
\bauthor{\bsnm{Xu}, \binits{C.}},
\bauthor{\bsnm{Liu}, \binits{J.}},
\bauthor{\bsnm{Tian}, \binits{Q.}}:
\bctitle{Unsupervised image super-resolution with an indirect supervised path}.
In: \bbtitle{Proceedings of the IEEE/CVF Conference on Computer Vision and
  Pattern Recognition Workshops},
pp. \bfpage{468}--\blpage{469}
(\byear{2020})
\end{bchapter}
\endbibitem

\bibitem{levin}
\begin{bchapter}
\bauthor{\bsnm{Levin}, \binits{A.}},
\bauthor{\bsnm{Weiss}, \binits{Y.}},
\bauthor{\bsnm{Durand}, \binits{F.}},
\bauthor{\bsnm{Freeman}, \binits{W.T.}}:
\bctitle{Understanding and evaluating blind deconvolution algorithms}.
In: \bbtitle{Proceedings of the IEEE/CVF Conference on Computer Vision and
  Pattern Recognition},
pp. \bfpage{1964}--\blpage{1971}
(\byear{2009}).
\bcomment{IEEE}
\end{bchapter}
\endbibitem

\bibitem{levin2011efficient}
\begin{bchapter}
\bauthor{\bsnm{{Levin, Anat and Weiss, Yair and Durand, Fredo and Freeman,
  William T}}}:
\bctitle{Efficient marginal likelihood optimization in blind deconvolution}.
In: \bbtitle{Proceedings of the IEEE/CVF Conference on Computer Vision and
  Pattern Recognition},
pp. \bfpage{2657}--\blpage{2664}
(\byear{2011}).
\bcomment{IEEE}
\end{bchapter}
\endbibitem

\bibitem{gan}
\begin{bchapter}
\bauthor{\bsnm{Goodfellow}, \binits{I.J.}},
\bauthor{\bsnm{Pouget-Abadie}, \binits{J.}},
\bauthor{\bsnm{Mirza}, \binits{M.}},
\bauthor{\bsnm{Xu}, \binits{B.}},
\bauthor{\bsnm{Warde-Farley}, \binits{D.}},
\bauthor{\bsnm{Ozair}, \binits{S.}},
\bauthor{\bsnm{Courville}, \binits{A.C.}},
\bauthor{\bsnm{Bengio}, \binits{Y.}}:
\bctitle{Generative adversarial nets}.
In: \bbtitle{Advances in Neural Information Processing Systems}
(\byear{2014})
\end{bchapter}
\endbibitem

\bibitem{pan}
\begin{barticle}
\bauthor{\bsnm{Pan}, \binits{J.}},
\bauthor{\bsnm{Sun}, \binits{D.}},
\bauthor{\bsnm{Pfister}, \binits{H.}},
\bauthor{\bsnm{Yang}, \binits{M.-H.}}:
\batitle{Deblurring images via dark channel prior}.
\bjtitle{IEEE Transactions on Pattern Analysis and Machine Intelligence}
\bvolume{40},
\bfpage{2315}--\blpage{2328}
(\byear{2018})
\end{barticle}
\endbibitem

\bibitem{pan2016blind}
\begin{bchapter}
\bauthor{\bsnm{{Pan, Jinshan and Sun, Deqing and Pfister, Hanspeter and Yang,
  Ming-Hsuan}}}:
\bctitle{Blind image deblurring using dark channel prior}.
In: \bbtitle{Proceedings of the IEEE/CVF Conference on Computer Vision and
  Pattern Recognition},
pp. \bfpage{1628}--\blpage{1636}
(\byear{2016})
\end{bchapter}
\endbibitem

\bibitem{extreme}
\begin{bchapter}
\bauthor{\bsnm{Yan}, \binits{Y.}},
\bauthor{\bsnm{Ren}, \binits{W.}},
\bauthor{\bsnm{Guo}, \binits{Y.}},
\bauthor{\bsnm{Wang}, \binits{R.}},
\bauthor{\bsnm{Cao}, \binits{X.}}:
\bctitle{Image deblurring via extreme channels prior}.
In: \bbtitle{Proceedings of the IEEE Conference on Computer Vision and Pattern
  Recognition},
pp. \bfpage{4003}--\blpage{4011}
(\byear{2017})
\end{bchapter}
\endbibitem

\bibitem{cai2020dark}
\begin{barticle}
\bauthor{\bsnm{Cai}, \binits{J.}},
\bauthor{\bsnm{Zuo}, \binits{W.}},
\bauthor{\bsnm{Zhang}, \binits{L.}}:
\batitle{Dark and bright channel prior embedded network for dynamic scene
  deblurring}.
\bjtitle{IEEE Transactions on Image Processing}
\bvolume{29},
\bfpage{6885}--\blpage{6897}
(\byear{2020})
\end{barticle}
\endbibitem

\bibitem{zssr_pre}
\begin{botherref}
\oauthor{\bsnm{Glasner}, \binits{D.}},
\oauthor{\bsnm{Bagon}, \binits{S.}},
\oauthor{\bsnm{Irani}, \binits{M.}}:
Super-resolution from a single image.
2009 IEEE 12th International Conference on Computer Vision,
349--356
(2009)
\end{botherref}
\endbibitem

\bibitem{zssr}
\begin{bchapter}
\bauthor{\bsnm{Shocher}, \binits{A.}},
\bauthor{\bsnm{Cohen}, \binits{N.}},
\bauthor{\bsnm{Irani}, \binits{M.}}:
\bctitle{"zero-shot" super-resolution using deep internal learning}.
In: \bbtitle{Proceedings of the IEEE/CVF Conference on Computer Vision and
  Pattern Recognition}
(\byear{2018})
\end{bchapter}
\endbibitem

\bibitem{mzsr}
\begin{bchapter}
\bauthor{\bsnm{Soh}, \binits{J.W.}},
\bauthor{\bsnm{Cho}, \binits{S.}},
\bauthor{\bsnm{Cho}, \binits{N.I.}}:
\bctitle{Meta-transfer learning for zero-shot super-resolution}.
In: \bbtitle{Proceedings of the IEEE/CVF Conference on Computer Vision and
  Pattern Recognition},
pp. \bfpage{3516}--\blpage{3525}
(\byear{2020})
\end{bchapter}
\endbibitem

\bibitem{srmd}
\begin{bchapter}
\bauthor{\bsnm{Zhang}, \binits{K.}},
\bauthor{\bsnm{Zuo}, \binits{W.}},
\bauthor{\bsnm{Zhang}, \binits{L.}}:
\bctitle{Learning a single convolutional super-resolution network for multiple
  degradations}.
In: \bbtitle{Proceedings of the IEEE/CVF Conference on Computer Vision and
  Pattern Recognition},
pp. \bfpage{3262}--\blpage{3271}
(\byear{2018})
\end{bchapter}
\endbibitem

\bibitem{ikc}
\begin{bchapter}
\bauthor{\bsnm{Gu}, \binits{J.}},
\bauthor{\bsnm{Lu}, \binits{H.}},
\bauthor{\bsnm{Zuo}, \binits{W.}},
\bauthor{\bsnm{Dong}, \binits{C.}}:
\bctitle{Blind super-resolution with iterative kernel correction}.
In: \bbtitle{Proceedings of the IEEE/CVF Conference on Computer Vision and
  Pattern Recognition},
pp. \bfpage{1604}--\blpage{1613}
(\byear{2019})
\end{bchapter}
\endbibitem

\bibitem{koalanet}
\begin{bchapter}
\bauthor{\bsnm{Kim}, \binits{S.Y.}},
\bauthor{\bsnm{Sim}, \binits{H.}},
\bauthor{\bsnm{Kim}, \binits{M.}}:
\bctitle{Koalanet: Blind super-resolution using kernel-oriented adaptive local
  adjustment}.
In: \bbtitle{Proceedings of the IEEE/CVF Conference on Computer Vision and
  Pattern Recognition},
pp. \bfpage{10611}--\blpage{10620}
(\byear{2021})
\end{bchapter}
\endbibitem

\bibitem{dipfkp}
\begin{bchapter}
\bauthor{\bsnm{Liang}, \binits{J.}},
\bauthor{\bsnm{Zhang}, \binits{K.}},
\bauthor{\bsnm{Gu}, \binits{S.}},
\bauthor{\bsnm{Van~Gool}, \binits{L.}},
\bauthor{\bsnm{Timofte}, \binits{R.}}:
\bctitle{Flow-based kernel prior with application to blind super-resolution}.
In: \bbtitle{Proceedings of the IEEE/CVF Conference on Computer Vision and
  Pattern Recognition},
pp. \bfpage{10601}--\blpage{10610}
(\byear{2021})
\end{bchapter}
\endbibitem

\bibitem{dasr}
\begin{bchapter}
\bauthor{\bsnm{Wei}, \binits{Y.}},
\bauthor{\bsnm{Gu}, \binits{S.}},
\bauthor{\bsnm{Li}, \binits{Y.}},
\bauthor{\bsnm{Timofte}, \binits{R.}},
\bauthor{\bsnm{Jin}, \binits{L.}},
\bauthor{\bsnm{Song}, \binits{H.}}:
\bctitle{Unsupervised real-world image super resolution via domain-distance
  aware training}.
In: \bbtitle{Proceedings of the IEEE/CVF Conference on Computer Vision and
  Pattern Recognition},
pp. \bfpage{13385}--\blpage{13394}
(\byear{2021})
\end{bchapter}
\endbibitem

\bibitem{real-esrgan}
\begin{botherref}
\oauthor{\bsnm{Wang}, \binits{X.}},
\oauthor{\bsnm{Xie}, \binits{L.}},
\oauthor{\bsnm{Dong}, \binits{C.}},
\oauthor{\bsnm{Shan}, \binits{Y.}}:
Real-esrgan: Training real-world blind super-resolution with pure synthetic
  data.
arXiv preprint arXiv:2107.10833
(2021)
\end{botherref}
\endbibitem

\bibitem{bsrgan}
\begin{botherref}
\oauthor{\bsnm{Zhang}, \binits{K.}},
\oauthor{\bsnm{Liang}, \binits{J.}},
\oauthor{\bsnm{Van~Gool}, \binits{L.}},
\oauthor{\bsnm{Timofte}, \binits{R.}}:
Designing a practical degradation model for deep blind image super-resolution.
arXiv preprint arXiv:2103.14006
(2021)
\end{botherref}
\endbibitem

\bibitem{inter_clr}
\begin{botherref}
\oauthor{\bsnm{Xie}, \binits{J.}},
\oauthor{\bsnm{Zhan}, \binits{X.}},
\oauthor{\bsnm{Liu}, \binits{Z.}},
\oauthor{\bsnm{Ong}, \binits{Y.-S.}},
\oauthor{\bsnm{Loy}, \binits{C.C.}}:
Delving into inter-image invariance for unsupervised visual representations.
International Journal of Computer Vision,
1--20
(2022)
\end{botherref}
\endbibitem

\bibitem{cdc}
\begin{bchapter}
\bauthor{\bsnm{Wei}, \binits{P.}},
\bauthor{\bsnm{Xie}, \binits{Z.}},
\bauthor{\bsnm{Lu}, \binits{H.}},
\bauthor{\bsnm{Zhan}, \binits{Z.}},
\bauthor{\bsnm{Ye}, \binits{Q.}},
\bauthor{\bsnm{Zuo}, \binits{W.}},
\bauthor{\bsnm{Lin}, \binits{L.}}:
\bctitle{Component divide-and-conquer for real-world image super-resolution}.
In: \bbtitle{European Conference on Computer Vision},
pp. \bfpage{101}--\blpage{117}
(\byear{2020}).
\bcomment{Springer}
\end{bchapter}
\endbibitem

\bibitem{lp-kpn}
\begin{bchapter}
\bauthor{\bsnm{Cai}, \binits{J.}},
\bauthor{\bsnm{Zeng}, \binits{H.}},
\bauthor{\bsnm{Yong}, \binits{H.}},
\bauthor{\bsnm{Cao}, \binits{Z.}},
\bauthor{\bsnm{Zhang}, \binits{L.}}:
\bctitle{Toward real-world single image super-resolution: A new benchmark and a
  new model}.
In: \bbtitle{Proceedings of the IEEE/CVF International Conference on Computer
  Vision},
pp. \bfpage{3086}--\blpage{3095}
(\byear{2019})
\end{bchapter}
\endbibitem

\bibitem{zhang2017learning}
\begin{bchapter}
\bauthor{\bsnm{Zhang}, \binits{K.}},
\bauthor{\bsnm{Zuo}, \binits{W.}},
\bauthor{\bsnm{Gu}, \binits{S.}},
\bauthor{\bsnm{Zhang}, \binits{L.}}:
\bctitle{Learning deep cnn denoiser prior for image restoration}.
In: \bbtitle{Proceedings of the IEEE/CVF Conference on Computer Cision and
  Pattern Recognition},
pp. \bfpage{3929}--\blpage{3938}
(\byear{2017})
\end{bchapter}
\endbibitem

\bibitem{selfdeblur}
\begin{bchapter}
\bauthor{\bsnm{Ren}, \binits{D.}},
\bauthor{\bsnm{Zhang}, \binits{K.}},
\bauthor{\bsnm{Wang}, \binits{Q.}},
\bauthor{\bsnm{Hu}, \binits{Q.}},
\bauthor{\bsnm{Zuo}, \binits{W.}}:
\bctitle{Neural blind deconvolution using deep priors}.
In: \bbtitle{Proceedings of the IEEE/CVF Conference on Computer Vision and
  Pattern Recognition},
pp. \bfpage{3341}--\blpage{3350}
(\byear{2020})
\end{bchapter}
\endbibitem

\bibitem{deblurpair}
\begin{bchapter}
\bauthor{\bsnm{Kaufman}, \binits{A.}},
\bauthor{\bsnm{Fattal}, \binits{R.}}:
\bctitle{Deblurring using analysis-synthesis networks pair}.
In: \bbtitle{Proceedings of the IEEE/CVF Conference on Computer Vision and
  Pattern Recognition},
pp. \bfpage{5811}--\blpage{5820}
(\byear{2020})
\end{bchapter}
\endbibitem

\bibitem{generalized_gauss}
\begin{barticle}
\bauthor{\bsnm{Pascal}, \binits{F.}},
\bauthor{\bsnm{Bombrun}, \binits{L.}},
\bauthor{\bsnm{Tourneret}, \binits{J.-Y.}},
\bauthor{\bsnm{Berthoumieu}, \binits{Y.}}:
\batitle{Parameter estimation for multivariate generalized gaussian
  distributions}.
\bjtitle{IEEE Transactions on Signal Processing}
\bvolume{61}(\bissue{23}),
\bfpage{5960}--\blpage{5971}
(\byear{2013})
\end{barticle}
\endbibitem

\bibitem{unprocessing_raw}
\begin{bchapter}
\bauthor{\bsnm{Brooks}, \binits{T.}},
\bauthor{\bsnm{Mildenhall}, \binits{B.}},
\bauthor{\bsnm{Xue}, \binits{T.}},
\bauthor{\bsnm{Chen}, \binits{J.}},
\bauthor{\bsnm{Sharlet}, \binits{D.}},
\bauthor{\bsnm{Barron}, \binits{J.T.}}:
\bctitle{Unprocessing images for learned raw denoising}.
In: \bbtitle{Proceedings of the IEEE/CVF Conference on Computer Vision and
  Pattern Recognition},
pp. \bfpage{11036}--\blpage{11045}
(\byear{2019})
\end{bchapter}
\endbibitem

\bibitem{jpeg}
\begin{bchapter}
\bauthor{\bsnm{Shin}, \binits{R.}},
\bauthor{\bsnm{Song}, \binits{D.}}:
\bctitle{Jpeg-resistant adversarial images}.
In: \bbtitle{NIPS 2017 Workshop on Machine Learning and Computer Security},
vol. \bseriesno{1}
(\byear{2017})
\end{bchapter}
\endbibitem

\bibitem{edsr}
\begin{bchapter}
\bauthor{\bsnm{Lim}, \binits{B.}},
\bauthor{\bsnm{Son}, \binits{S.}},
\bauthor{\bsnm{Kim}, \binits{H.}},
\bauthor{\bsnm{Nah}, \binits{S.}},
\bauthor{\bsnm{Mu~Lee}, \binits{K.}}:
\bctitle{Enhanced deep residual networks for single image super-resolution}.
In: \bbtitle{Proceedings of the IEEE/CVF Conference on Computer Vision and
  Pattern Recognition Workshops},
pp. \bfpage{136}--\blpage{144}
(\byear{2017})
\end{bchapter}
\endbibitem

\bibitem{adatarget}
\begin{bchapter}
\bauthor{\bsnm{Jo}, \binits{Y.}},
\bauthor{\bsnm{Oh}, \binits{S.W.}},
\bauthor{\bsnm{Vajda}, \binits{P.}},
\bauthor{\bsnm{Kim}, \binits{S.J.}}:
\bctitle{Tackling the ill-posedness of super-resolution through adaptive target
  generation}.
In: \bbtitle{Proceedings of the IEEE/CVF Conference on Computer Vision and
  Pattern Recognition},
pp. \bfpage{16236}--\blpage{16245}
(\byear{2021})
\end{bchapter}
\endbibitem

\bibitem{div2k}
\begin{botherref}
\oauthor{\bsnm{Agustsson}, \binits{E.}},
\oauthor{\bsnm{Timofte}, \binits{R.}}:
Ntire 2017 challenge on single image super-resolution: Dataset and study,
1122--1131
(2017)
\end{botherref}
\endbibitem

\bibitem{flickr2k}
\begin{botherref}
\oauthor{\bparticle{et} \bsnm{al.}, \binits{R.T.}}:
Ntire 2017 challenge on single image super-resolution: Methods and results,
1110--1121
(2017)
\end{botherref}
\endbibitem

\bibitem{ntire2020}
\begin{bchapter}
\bauthor{\bsnm{Lugmayr}, \binits{A.}},
\bauthor{\bsnm{Danelljan}, \binits{M.}},
\bauthor{\bsnm{Timofte}, \binits{R.}}:
\bctitle{Ntire 2020 challenge on real-world image super-resolution: Methods and
  results}.
In: \bbtitle{Proceedings of the IEEE/CVF Conference on Computer Vision and
  Pattern Recognition Workshops},
pp. \bfpage{494}--\blpage{495}
(\byear{2020})
\end{bchapter}
\endbibitem

\bibitem{dped}
\begin{bchapter}
\bauthor{\bsnm{Ignatov}, \binits{A.}},
\bauthor{\bsnm{Kobyshev}, \binits{N.}},
\bauthor{\bsnm{Timofte}, \binits{R.}},
\bauthor{\bsnm{Vanhoey}, \binits{K.}},
\bauthor{\bsnm{Van~Gool}, \binits{L.}}:
\bctitle{Dslr-quality photos on mobile devices with deep convolutional
  networks}.
In: \bbtitle{Proceedings of the IEEE International Conference on Computer
  Vision},
pp. \bfpage{3277}--\blpage{3285}
(\byear{2017})
\end{bchapter}
\endbibitem

\bibitem{martin2001database}
\begin{bchapter}
\bauthor{\bsnm{Martin}, \binits{D.}},
\bauthor{\bsnm{Fowlkes}, \binits{C.}},
\bauthor{\bsnm{Tal}, \binits{D.}},
\bauthor{\bsnm{Malik}, \binits{J.}}:
\bctitle{A database of human segmented natural images and its application to
  evaluating segmentation algorithms and measuring ecological statistics}.
In: \bbtitle{Proceedings Eighth IEEE International Conference on Computer
  Vision. ICCV 2001},
vol. \bseriesno{2},
pp. \bfpage{416}--\blpage{423}
(\byear{2001}).
\bcomment{IEEE}
\end{bchapter}
\endbibitem

\bibitem{manga109}
\begin{barticle}
\bauthor{\bsnm{Matsui}, \binits{Y.}},
\bauthor{\bsnm{Ito}, \binits{K.}},
\bauthor{\bsnm{Aramaki}, \binits{Y.}},
\bauthor{\bsnm{Fujimoto}, \binits{A.}},
\bauthor{\bsnm{Ogawa}, \binits{T.}},
\bauthor{\bsnm{Yamasaki}, \binits{T.}},
\bauthor{\bsnm{Aizawa}, \binits{K.}}:
\batitle{Sketch-based manga retrieval using manga109 dataset}.
\bjtitle{Multimedia Tools and Applications}
\bvolume{76},
\bfpage{21811}--\blpage{21838}
(\byear{2016})
\end{barticle}
\endbibitem

\bibitem{ffdnet}
\begin{barticle}
\bauthor{\bsnm{Zhang}, \binits{K.}},
\bauthor{\bsnm{Zuo}, \binits{W.}},
\bauthor{\bsnm{Zhang}, \binits{L.}}:
\batitle{Ffdnet: Toward a fast and flexible solution for cnn-based image
  denoising}.
\bjtitle{IEEE Transactions on Image Processing}
\bvolume{27}(\bissue{9}),
\bfpage{4608}--\blpage{4622}
(\byear{2018})
\end{barticle}
\endbibitem

\bibitem{ssim}
\begin{barticle}
\bauthor{\bsnm{Wang}, \binits{Z.}},
\bauthor{\bsnm{Bovik}, \binits{A.C.}},
\bauthor{\bsnm{Sheikh}, \binits{H.R.}},
\bauthor{\bsnm{Simoncelli}, \binits{E.P.}}:
\batitle{Image quality assessment: from error visibility to structural
  similarity}.
\bjtitle{IEEE transactions on image processing}
\bvolume{13}(\bissue{4}),
\bfpage{600}--\blpage{612}
(\byear{2004})
\end{barticle}
\endbibitem

\bibitem{lpips}
\begin{bchapter}
\bauthor{\bsnm{Zhang}, \binits{R.}},
\bauthor{\bsnm{Isola}, \binits{P.}},
\bauthor{\bsnm{Efros}, \binits{A.A.}},
\bauthor{\bsnm{Shechtman}, \binits{E.}},
\bauthor{\bsnm{Wang}, \binits{O.}}:
\bctitle{The unreasonable effectiveness of deep features as a perceptual
  metric}.
In: \bbtitle{Proceedings of the IEEE/CVF Conference on Computer Vision and
  Pattern Recognition}
(\byear{2018})
\end{bchapter}
\endbibitem

\bibitem{alex}
\begin{barticle}
\bauthor{\bsnm{Krizhevsky}, \binits{A.}},
\bauthor{\bsnm{Sutskever}, \binits{I.}},
\bauthor{\bsnm{Hinton}, \binits{G.E.}}:
\batitle{Imagenet classification with deep convolutional neural networks}.
\bjtitle{Advances in Neural Information Processing Systems}
\bvolume{25},
\bfpage{1097}--\blpage{1105}
(\byear{2012})
\end{barticle}
\endbibitem

\bibitem{niqe}
\begin{barticle}
\bauthor{\bsnm{Mittal}, \binits{A.}},
\bauthor{\bsnm{Soundararajan}, \binits{R.}},
\bauthor{\bsnm{Bovik}, \binits{A.C.}}:
\batitle{Making a “completely blind” image quality analyzer}.
\bjtitle{IEEE Signal Processing Letters}
\bvolume{20}(\bissue{3}),
\bfpage{209}--\blpage{212}
(\byear{2012})
\end{barticle}
\endbibitem

\bibitem{nrqm}
\begin{barticle}
\bauthor{\bsnm{Ma}, \binits{C.}},
\bauthor{\bsnm{Yang}, \binits{C.-Y.}},
\bauthor{\bsnm{Yang}, \binits{X.}},
\bauthor{\bsnm{Yang}, \binits{M.-H.}}:
\batitle{Learning a no-reference quality metric for single-image
  super-resolution}.
\bjtitle{Computer Vision and Image Understanding}
\bvolume{158},
\bfpage{1}--\blpage{16}
(\byear{2017})
\end{barticle}
\endbibitem

\bibitem{pi}
\begin{bchapter}
\bauthor{\bsnm{Blau}, \binits{Y.}},
\bauthor{\bsnm{Mechrez}, \binits{R.}},
\bauthor{\bsnm{Timofte}, \binits{R.}},
\bauthor{\bsnm{Michaeli}, \binits{T.}},
\bauthor{\bsnm{Zelnik-Manor}, \binits{L.}}:
\bctitle{The 2018 pirm challenge on perceptual image super-resolution}.
In: \bbtitle{Proceedings of the European Conference on Computer Vision (ECCV)
  Workshops},
pp. \bfpage{0}--\blpage{0}
(\byear{2018})
\end{bchapter}
\endbibitem

\bibitem{srgan}
\begin{bchapter}
\bauthor{\bsnm{Ledig}, \binits{C.}},
\bauthor{\bsnm{Theis}, \binits{L.}},
\bauthor{\bsnm{Husz{\'a}r}, \binits{F.}},
\bauthor{\bsnm{Caballero}, \binits{J.}},
\bauthor{\bsnm{Cunningham}, \binits{A.}},
\bauthor{\bsnm{Acosta}, \binits{A.}},
\bauthor{\bsnm{Aitken}, \binits{A.}},
\bauthor{\bsnm{Tejani}, \binits{A.}},
\bauthor{\bsnm{Totz}, \binits{J.}},
\bauthor{\bsnm{Wang}, \binits{Z.}}, \betal:
\bctitle{Photo-realistic single image super-resolution using a generative
  adversarial network}.
In: \bbtitle{Proceedings of the IEEE/CVF Conference on Computer Vision and
  Pattern Recognition},
pp. \bfpage{4681}--\blpage{4690}
(\byear{2017})
\end{bchapter}
\endbibitem

\bibitem{vgg}
\begin{botherref}
\oauthor{\bsnm{Simonyan}, \binits{K.}},
\oauthor{\bsnm{Zisserman}, \binits{A.}}:
Very deep convolutional networks for large-scale image recognition.
arXiv preprint arXiv:1409.1556
(2014)
\end{botherref}
\endbibitem

\bibitem{adam}
\begin{botherref}
\oauthor{\bsnm{Kingma}, \binits{D.P.}},
\oauthor{\bsnm{Ba}, \binits{J.}}:
Adam: A method for stochastic optimization.
arXiv preprint arXiv:1412.6980
(2014)
\end{botherref}
\endbibitem

\bibitem{ji2020real}
\begin{bchapter}
\bauthor{\bsnm{Ji}, \binits{X.}},
\bauthor{\bsnm{Cao}, \binits{Y.}},
\bauthor{\bsnm{Tai}, \binits{Y.}},
\bauthor{\bsnm{Wang}, \binits{C.}},
\bauthor{\bsnm{Li}, \binits{J.}},
\bauthor{\bsnm{Huang}, \binits{F.}}:
\bctitle{Real-world super-resolution via kernel estimation and noise
  injection}.
In: \bbtitle{Proceedings of the IEEE/CVF Conference on Computer Vision and
  Pattern Recognition Workshops},
pp. \bfpage{466}--\blpage{467}
(\byear{2020})
\end{bchapter}
\endbibitem

\bibitem{pdmsr}
\begin{bchapter}
\bauthor{\bsnm{Luo}, \binits{Z.}},
\bauthor{\bsnm{Huang}, \binits{Y.}},
\bauthor{},
\bauthor{\bsnm{Li}, \binits{S.}},
\bauthor{\bsnm{Wang}, \binits{L.}},
\bauthor{\bsnm{Tan}, \binits{T.}}:
\bctitle{Learning the degradation distribution for blind image
  super-resolution}.
In: \bbtitle{Proceedings of the IEEE/CVF Conference on Computer Vision and
  Pattern Recognition}
(\byear{2022})
\end{bchapter}
\endbibitem

\bibitem{corrfilter}
\begin{bchapter}
\bauthor{\bsnm{Hussein}, \binits{S.A.}},
\bauthor{\bsnm{Tirer}, \binits{T.}},
\bauthor{\bsnm{Giryes}, \binits{R.}}:
\bctitle{Correction filter for single image super-resolution: Robustifying
  off-the-shelf deep super-resolvers}.
In: \bbtitle{Proceedings of the IEEE/CVF Conference on Computer Vision and
  Pattern Recognition},
pp. \bfpage{1428}--\blpage{1437}
(\byear{2020})
\end{bchapter}
\endbibitem

\bibitem{pan2014}
\begin{bchapter}
\bauthor{\bsnm{Pan}, \binits{J.}},
\bauthor{\bsnm{Hu}, \binits{Z.}},
\bauthor{\bsnm{Su}, \binits{Z.}},
\bauthor{\bsnm{Yang}, \binits{M.-H.}}:
\bctitle{Deblurring text images via l0-regularized intensity and gradient
  prior}.
In: \bbtitle{Proceedings of the IEEE/CVF Conference on Computer Vision and
  Pattern Recognition},
pp. \bfpage{2901}--\blpage{2908}
(\byear{2014})
\end{bchapter}
\endbibitem

\end{thebibliography}
}


\end{document}